\documentclass[runningheads]{llncs}

\usepackage[final,year=2026,ID=423]{eccv}

\usepackage{eccvabbrv}

\usepackage{graphicx}
\usepackage{booktabs}

\usepackage[accsupp]{axessibility}  

\usepackage{url}            
\usepackage{booktabs}       
\usepackage{amsfonts}       
\usepackage{nicefrac}       
\usepackage{microtype}      
\usepackage{graphicx}
\usepackage{booktabs}
\usepackage{amsmath}
\usepackage{amssymb}
\usepackage{booktabs}
\usepackage{verbatim}
\usepackage{multicol}
\usepackage{multirow}
\usepackage{enumitem}
\usepackage{overpic}
\usepackage{adjustbox}
\usepackage{subcaption}
\newcommand{\myPara}[1]{\vspace{2pt}\noindent\textbf{#1}}

\usepackage{makecell}
\usepackage{xspace}
\newcommand{\nameofmethod}{GeoWorld}
\usepackage{wrapfig}

\usepackage[breaklinks,colorlinks,citecolor=eccvblue]{hyperref}

\usepackage{orcidlink}
\usepackage{marvosym}

\begin{document}

\title{GeoWorld: Providing Full-frame Geometry Features to Facilitate 3D Scene Generation}

\titlerunning{GeoWorld, ECCV 2026}

\author{Yuhao Wan\inst{1,2}\orcidlink{0000-0001-7526-5247} \and Lijuan Liu\inst{2} \and Jingzhi Zhou\inst{1} \and Zihan Zhou\inst{3} \and Xuying Zhang\inst{1} \and Dongbo Zhang\inst{2} \and Shaohui Jiao\inst{2} \and Qibin Hou\inst{1,4}\orcidlink{0000-0002-8388-8708} \and Ming-Ming Cheng\inst{4,1}\orcidlink{0000-0001-5550-8758}\Letter} 

\authorrunning{Wan et al.}

\institute{$^1$VCIP \& AAIS, Nankai University \qquad 
$^2$ByteDance Inc. \\
$^3$Renmin University of China \qquad
$^4$NKIARI, Shenzhen Futian }

\maketitle
\begin{center}
  \vspace{-0.3cm}
  \setlength{\abovecaptionskip}{2pt}
  \includegraphics[width=\textwidth]{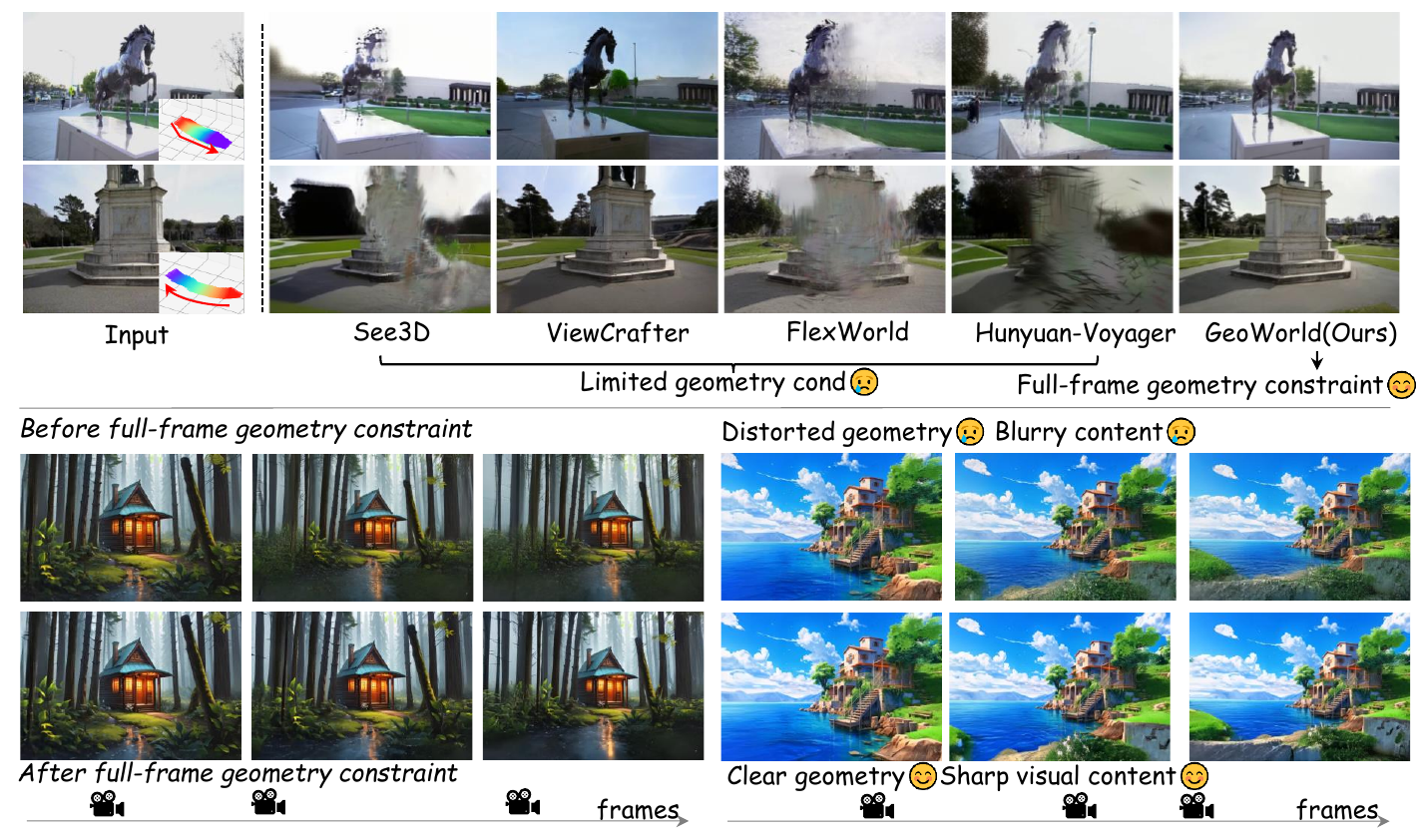}
  \captionof{figure}{
   Visual comparisons. Top: Comparison between our \textbf{GeoWorld} and previous methods. 
   By incorporating full-frame geometry constraints, our approach achieves superior visual quality. 
   Bottom left: Results before applying full-frame geometry constraints, which often suffer from geometric distortions and blurry content. 
   Bottom right: Results after applying full-frame geometry constraints. 
   By unlocking the potential of geometry models, our \textbf{GeoWorld} produces clear geometric structures and sharp visual details.
  }\label{fig:teaser}
\end{center}

\begin{abstract}

Previous works that leverage video models for image-to-3D scene generation often suffer from geometric distortions and blurry content. 
Using video generation models to implicitly maintain geometric consistency according to a single-frame input is ineffective. 
In this paper, we present a two-stage method, named \textbf{GeoWorld}, that renovates the image-to-3D scene generation pipeline by providing full-frame geometry features. 
The first-stage video generation model, followed by a multi-view geometry model, produces \textbf{full-frame} geometry features,
which are then used as a mental draft of geometric conditions to aid the second-stage video-generation model.
A geometric loss is proposed to impose real-world geometric constraints, and a geometry adaptation module is introduced to ensure the effective utilization of geometry features.
Thanks to full-frame geometric modeling, the two smaller video models in our two-stage method can generate higher-fidelity 3D scenes than SOTA methods, while being even faster, \eg $7.5 \times$ faster than Hunyuan-Voyager.
Project page: \url{https://peaes.github.io/GeoWorld}.
\end{abstract}

\section{Introduction}
\label{sec:Introduction}
Generating a high-fidelity 3D scene from a single image has become a significant topic in recent years due to its high value in applications such as entertainment, interior and architectural design, and autonomous driving~\cite{cat3d, Magicdrive, Uniscene, Dist-4d, AR-1-to-3, SEVA, Cast, wang2025diffusion, zhou2025onevae}.
Leveraging deep learning methods for this task can significantly advance traditional 3D modeling pipelines.
Given the limited information in a single image, a common approach is to use generative model priors to synthesize the scene content.
Early methods~\cite{SceneWiz3D, Dreamfusion, Set-the-scene, Dreamscene360, Dreamscene, Luciddreamer, Wonderworld, Text2room} often rely on 2D generative models~\cite{LDM, DDPM}, which often lead to issues such as structural inconsistency and inconsistencies within the scene content.

\begin{figure}[t]
  \centering
  \includegraphics[width=\linewidth]{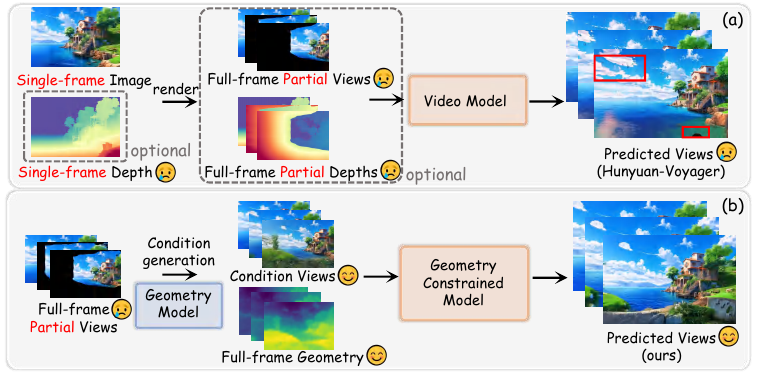}
  \caption{Pipeline comparison. 
    (a) Pipelines of previous methods~\cite{Voyager, FlexWorld, Viewcrafter}. 
    Although details vary, their video models are conditioned only on single-frame information and limited geometric information. 
    `Optional' indicates that this step is not included in some methods. 
    We can see that partial views or partial depths have very limited information. 
    (b) Our GeoWorld leverages the geometric condition generation procedure and a geometry model to obtain full-frame geometry features for generation, rather than relying solely on geometry extracted from the input image.
  }\label{fig:intro}
\end{figure}

Thanks to the advances in foundational 3D generative models, some recent works~\cite{Viewcrafter, Dimensionx, FlexWorld, Voyager, Stargen, IDCNet, flashworld, Matrixgame, yan, Motionstream, Yume, Magicworld, InfiniteWorld, Worldplay, Gen3R} use video models~\cite{Cogvideo, Cogvideox, Wan} and leverage their implicit 3D priors to alleviate the aforementioned issues.
Such methods typically employ video models to synthesize a video under a specified camera trajectory from a single input image, and subsequently reconstruct a 3D scene from the generated video. 
However, generating high-fidelity videos from a single image remains challenging. 
As shown at the top of Fig.~\ref{fig:teaser}, these methods often suffer from geometric distortions and blurry content, which degrade the quality of the final 3D reconstruction.
To address the above issues, a common approach is to provide additional geometric guidance for the model. 
As shown in Fig.~\ref{fig:intro}(a), some previous works~\cite{Voyager, FlexWorld, Viewcrafter} have utilized estimated monocular depth maps as spatial priors or camera embeddings to assist in video generation. `Optional' indicates that this step is not included in some methods. %

Compared to tasks such as 3D reconstruction~\cite{VGGT, Worldmirror, da3} or multi-view-to-3D scene generation~\cite{Stargen, cat3d}, image-to-3D scene generation is significantly more ill-posed. 
How to effectively provide geometric guidance for this task remains an open question. 
Previous works rely on limited geometric information extracted from a single input frame, which is insufficient to guide the generation of an entire video. 
Consequently, even larger models struggle to produce satisfactory results~\cite{FlexWorld, Voyager}.
However, providing corresponding geometric information for every frame is extremely challenging. 
Fortunately, we experimentally find that feeding rendered partial views into the video model produces coarse but content-complete condition views, as shown in Fig.~\ref{fig:intro}(b). 
These views can be used to extract geometric information, thereby providing full-frame guidance for the subsequent generation process.
Inspired by this, we propose a new two-stage pipeline paradigm for this task that incorporates full-frame geometric features into the video generation process. 
As shown in Fig.~\ref{fig:intro}(b), in the first stage, we generate a video with complete content, and then leverage geometry models that can extract geometric information from multiple frames to provide geometry features. 
These features are then used to facilitate the second-stage video generation model.

To be specific, we first render the single-frame input using the given camera trajectory.
The resulting rendered partial views are fed into a fine-tuned video model to generate a coarse but content-complete video. 
This video can then be used as input to a geometry model to extract full-frame geometry features, providing a mental draft of geometric conditions for the second-stage video generation model.
Since the geometric information is derived from a model of limited accuracy, we further design a geometric alignment loss to compensate for this limitation.
Instead of directly embedding the obtained geometry features as conditioning, our geometric alignment loss aligns the geometry features extracted from the predicted and ground-truth videos during training. 
This design imposes real-world geometric constraints on the model, and the number of frames is naturally consistent.
Finally, we propose a geometry adaptation module to effectively exploit the extracted full-frame geometry features for improving video generation quality.
From Fig.~\ref{fig:teaser}, we can see that our \nameofmethod{} is able to generate a high-fidelity 3D scene from a single image and a given camera trajectory, outperforming other state-of-the-art methods.
Counterintuitively, although we introduce an additional procedure for obtaining full-frame geometry features, our two-stage method is more efficient, \eg uses only $0.3 \times$ the model size and achieves $7.5 \times$ faster inference than Hunyuan-Voyager~\cite{Voyager} (See Sec.~\ref{sec:ca}).
%
Our contributions can be summarized as follows:
\begin{itemize}[leftmargin=*]
\item We propose GeoWorld, a novel two-stage pipeline paradigm for single-image-to-3D scene generation. 
We incorporate full-frame geometric features into the video generation process to alleviate the difficulty of generation caused by the limited input information inherent in this task.
\item We explore how to leverage geometry models to assist the video generation process. 
In this exploration, we design a geometric condition generation procedure, a geometric alignment loss, and a geometry adaptation module to gradually unlock the potential of geometric information.
\item Our \nameofmethod{} outperforms previous methods qualitatively, achieves superior fidelity, yet is more computationally efficient.
\end{itemize}

\begin{figure*}[t]
  \centering
  \setlength{\abovecaptionskip}{0pt}
  \includegraphics[width=\linewidth]{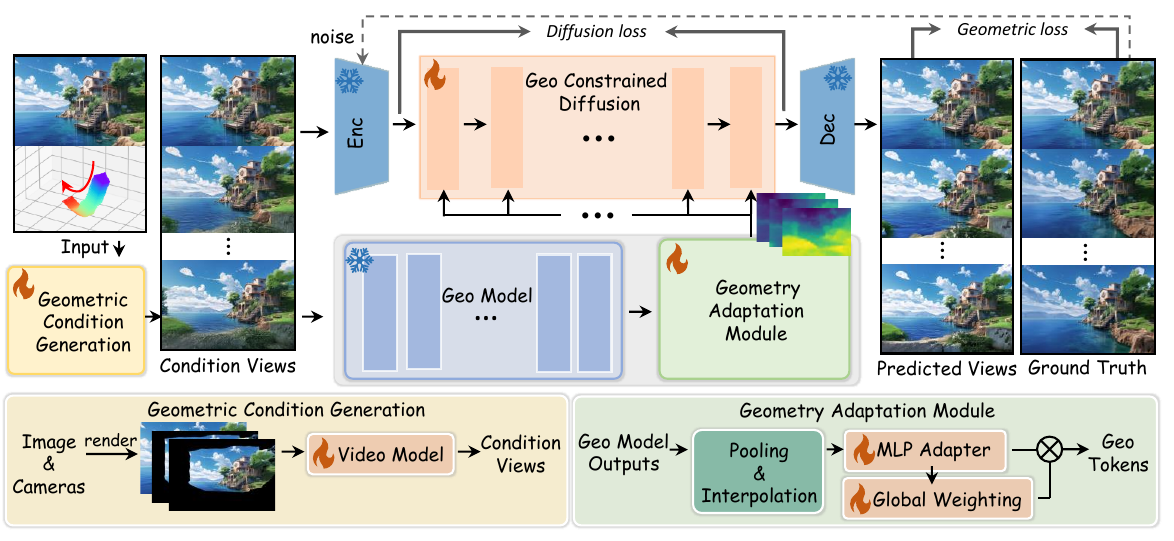}
  \caption{Overview of our \textbf{\nameofmethod{}}. 
    During training, we perform the geometric condition generation procedure, feeding the obtained condition views into the geometry model to obtain full-frame geometry features, which are then processed by the geometry adaptation module. 
    Together, the condition views and geometry features serve as input to the geometry-constrained diffusion model. 
    The predicted views produced by this model are then used to reconstruct the 3DGS scene.
  }\label{fig:arch}
\end{figure*}

\section{Related work}

\subsection{3D Scene Generation}

Existing scene generation methods can be broadly categorized into four types.
The first category comprises view-by-view image inpainting approaches~\cite{Rgbd2, Wonderjourney, Fastscene, Wonderworld, Luciddreamer, Dreamscene360, GenRC, Layerpano3d}, which generate frames sequentially along a predefined camera trajectory.
However, due to the lack of global semantic consistency, they struggle to produce semantically coherent full-scene results.
The second category directly generates 3D scene representations~\cite{DiffusionGS, Director3d, Gs-lrm, Flash3d, Wonderland} such as 3D Gaussian Splatting (3DGS)~\cite{3DGS}.
While these methods offer advantages in 3D consistency and reconstructability, their performance is often constrained by the scarcity of high-quality 3D data, posing challenges for training robust models.
The third category is compositional scene generation~\cite{HiScene, Midi, Architect, Cast, Gen3dsr}, where the key idea is to generate individual objects and place them plausibly within a scene layout.
Although significant progress has been made in 3D object generation, these methods still face unresolved issues related to model stability, object placement, and occlusion handling.
The final category is controllable video generation~\cite{See3D, Viewcrafter, Dimensionx, FlexWorld, IDCNet, Genxd, Cameractrl}, which aims to synthesize a sequence of spatially consistent video frames given a camera trajectory. This is typically achieved by fine-tuning pre-trained video diffusion models, which can produce visually appealing videos. However, such methods often exhibit geometric inconsistencies and structural artifacts that degrade the quality of subsequent 3D reconstructions.

To overcome these limitations, our method leverages multi-frame geometric information and introduces constraints to enforce spatial consistency, leading to highly competitive reconstruction results.

\subsection{Learning-Based 3D Reconstruction}
Unlike traditional 3D reconstruction pipelines that require training separately for each individual scene, learning-based reconstruction methods leverage neural networks trained on large-scale scene datasets to encode strong scene priors, ultimately achieving impressive open-world generalization capabilities~\cite{Cut3r, Dust3r, mast3r, VGGT, Fast3r, pi3, flare, da3, Dens3r, Worldmirror, meng20253d, wu2024recent, peng2025gaussian}. 
DUSt3R~\cite{Dust3r} directly regresses a 3D point cloud from a pair of RGB images. It extracts features from the two views using a transformer architecture enhanced with cross-attention mechanisms, and then feeds the fused features into a regression head to predict the point cloud along with a confidence map.
Its successor, MASt3R~\cite{mast3r}, maintains the two-view regression paradigm but introduces a confidence-weighted loss to improve prediction reliability.
Recent methods have generalized the two-view alignment-based architecture to handle multi-view scenarios, allowing for the joint processing of long frame sequences—up to 100 frames or more—as demonstrated by models such as Fast3R~\cite{Fast3r} and VGGT~\cite{VGGT}. VGGT scales up the core idea of DUSt3R into a 1.2B-parameter transformer that jointly predicts camera intrinsics and extrinsics, dense depth maps, 3D point clouds, and 2D tracking features. %
The features extracted by such architectures exhibit strong 3D consistency and can support a wide range of downstream 3D tasks.
Since video generation models often lack explicit 3D consistency control, introducing geometry features to guide the video generation process emerges as a natural and promising direction.

\section{Methodology}
\label{sec:Methodology}

As mentioned in Sec.~\ref{sec:Introduction}, our goal is to leverage geometry models to provide reliable full-frame conditional signals for the video generation process. In practice, we use VGGT~\cite{VGGT} as the geometry model. The overall architecture of our \nameofmethod{} is shown in Fig.~\ref{fig:arch}. Our design consists of three components: a geometric condition generation procedure to obtain full-frame geometry features, a geometric alignment loss to introduce real-world geometric constraints, and a geometry adaptation module to utilize the geometry features effectively. 
For the video generation process, during training, we first fine-tune a video model. We then let this model perform inference on the entire training set, and the newly generated data is used to train the geometry-constrained diffusion model. Finally, we use the predicted views to reconstruct the 3DGS scene.

\subsection{Geometric Condition Generation}
\label{sec:geopre}
The primary challenge of utilizing geometry models to help the video generation process lies in obtaining suitable geometry features. With only a single input frame, one can extract geometry for that frame alone, which provides limited guidance for subsequent frames in the video.
We solve this by introducing a geometric condition generation procedure to attain appropriate full-frame geometry features.
As shown in Fig.~\ref{fig:arch}, this procedure employs a fine-tuned video model to generate conditional views from the single-frame input and the given camera trajectory. In the full \nameofmethod{} pipeline, these conditional views are then processed by the geometry model to obtain full-frame geometry features. Specifically, the entire process consists of two components: rendering and completion.

\noindent\textbf{Rendering.} The goal of this component is to obtain a video, in which each frame is the rendering of the input single-frame image under the given camera trajectory. 
This video serves as the input to the video model. The rendering procedure differs between the training and inference phases to ensure higher-quality inputs during training. 
During training, we reconstruct the 3DGS scene using all available frames from the dataset, and then start from a random frame, extract its depth from the 3DGS, and perform back-projection and pairing process~\cite{FlexWorld}. During inference, we directly estimate the point cloud of the single-frame input under the given camera trajectory using MAST3R~\cite{mast3r}, and then perform back-projection.

\noindent\textbf{Completion.} The goal of this component is to use the fine-tuned video model to complete the rendered video into a content-complete one. Specifically, we first collect a batch of training data to fine-tune the video model directly. Then, the fine-tuned video model performs inference on the entire training dataset, and the newly generated data is used to train the geometry-constrained diffusion model shown in Fig.~\ref{fig:arch}. In terms of model design, since only single-frame geometry features are available at this stage, we simply use them as additional conditions and embed them into the model through cross-attention.

Through this process, we can obtain full-frame geometry features with the help of the geometry model. These outcomes also reflect, to some extent, the limitations of directly fine-tuning video models for geometrically consistent video generation. In Sec.~\ref{sec:ab}, we further discuss the quality of the conditional views (Fig.~\ref{fig:ab_geopre}) and validate the role of the geometry features in the subsequent optimization process through comparisons (Fig.~\ref{fig:ab_geoconstraints}).

\subsection{Geometric Alignment Loss}
\label{sec:geoloss}

Although the geometric condition generation process provides relatively complete geometric information, it is derived from a model of limited accuracy. To compensate for this limitation, we introduce real-world geometric information to guide the video generation process toward geometrically 3D scene synthesis. Specifically, we incorporate a geometric alignment loss into the diffusion objective, which compares the geometric features extracted from the generated and ground-truth videos. The loss is computed as the mean squared error between the two corresponding features obtained from the geometry model. In GeoWorld, this geometry model corresponds to the aggregator module of VGGT~\cite{VGGT}, with the decoding stage omitted to preserve complete geometric representations.

Specifically, given the ground-truth data $I$, a randomly sampled timestep $t$, and the corresponding Gaussian noise $\epsilon$, we first encode $I$ using the pre-trained VAE encoder and apply the forward process to obtain the input $\mathbf{x}$ for the geometry-constrained diffusion model. Geometry tokens $c_{geo}$ are obtained as described in Sec.~\ref{sec:geoadapt}. The diffusion loss is then defined as:
\begin{equation}
\mathcal{L}_{diff}=\mathbb{E}_{t, \epsilon\sim\mathcal{N}}\left[\left\| \epsilon - \epsilon_{\theta}(\mathbf{x}, c_{geo}, t) \right\|_2^2\right].
\end{equation}
Given the predicted views $I_{pred}$ and the geometry model $G$, the geometric loss is defined as:
\begin{equation}
\mathcal{L}_{geo}=\left\| G(I) - G(I_{pred}) \right\|_2^2.
\end{equation}
By leveraging the priors of the geometry model, this design could implicitly provide the model with real-world geometric information, ensuring that the optimization direction of the geometry-constrained model is geometrically consistent with the ground truth. The geometric alignment loss is defined as:
\begin{equation}
\mathcal{L}= \mathcal{L}_{diff} + \lambda \mathcal{L}_{geo},
\end{equation}
where $\lambda$ is used as the weight for the geometric loss. As shown in Fig.~\ref{fig:ab_modeldesign}, we observe that the outputs exhibit fewer geometric distortions and artifacts after incorporating the geometric alignment loss.

\subsection{Geometry Adaptation Module}
\label{sec:geoadapt}

The goal of this subsection is to enable the model to effectively utilize the geometry features. In the latent space, since the output $g$ of the geometry model differs from the input $\mathbf{x}$ of the video model along the frame, height, and width dimensions, we first perform pooling along the frame dimension and interpolation along the height and width dimensions to obtain $g_{resize}$, aligning its size with $\mathbf{x}$. We then train an MLP-based adapter to process the resized features and align them with the latent space of the video model, resulting in $g_{ada}$.
Since the generation of condition views lacks geometric guidance, some regions inevitably exhibit ambiguous geometric structures.
To prevent these ambiguous structures from misleading the model, we perform a global weighting on $g_{ada}$ after processing it with the MLP adapter.
We use an MLP-based predictor to integrate global information, similar to the Squeeze-and-Excitation block proposed in~\cite{SENet}, and output a global weight for each token.
By visualizing the global weights, we observe that low-weight tokens contain limited useful geometric information (Sec.~\ref{sec:ab}).
As a result, we multiply the global weights with $g_{ada}$ and empirically discard the bottom 50\% of tokens with lower weights to achieve a better performance, resulting in the final geometry tokens $c_{geo}$. %
Finally, we fuse the obtained $c_{geo}$ with $\mathbf{x}$ using a single-frame cross-attention in each layer. Given the query $Q$ from $\mathbf{x}$, key $K$ from $c_{geo}$, and value $V$ from $c_{geo}$, the formulation can be written as follows:
\begin{equation}
\operatorname{CA}(\mathbf{Q}, \mathbf{K}, \mathbf{V})= 
\operatorname{Softmax}\left(\frac{\mathbf{Q} \mathbf{K}^{T}}{\sqrt{d_{k}}} + \mathbf{B} \right) \mathbf{V},
\end{equation}
where $B$ is an aligned relative position embedding and $\sqrt{d_{k}}$ is a scaling factor~\cite{dosovitskiy2020image}.

\section{Experiments}

\subsection{Experimental Settings}

\myPara{Model and training details.}
We use Wan2.1-1.3B~\cite{Wan} as our video model. During training, we set the batch size, learning rate, input image resolution, video frame length, and the weight for the geometric loss to 16, 5e-5, 192$\times$336, 17, and 0.2, respectively.
The video model in the geometric condition generation procedure and the geometry-constrained model are trained for 7000 and 2000 iterations, respectively. Our geometry adaptation module and the geometry-constrained diffusion model are trained simultaneously. The training process is conducted on 8 NVIDIA A100 GPUs.

\myPara{Training Dataset.} We use DL3DV~\cite{Dl3dv} as our training dataset and construct training pairs following the dataset construction method proposed in FlexWorld~\cite{FlexWorld} (See Sec.~\ref{sec:geopre}).
Specifically, we sample two epochs from DL3DV and select the top 25\% of cases with the smallest average camera translation and rotation to ensure the quality of the partial views, resulting in approximately 5000 video pairs. We found that this simple filtering strategy effectively removes low-quality and facilitates training.

\begin{table*}[t]
\setlength{\abovecaptionskip}{2pt}
\setlength\tabcolsep{4pt}
\scriptsize
\centering
\caption{
Quantitative comparison of our \nameofmethod{} with recent state-of-the-art methods on \textbf{novel view synthesis}. The best performances are in \textbf{bold} and the second performances are \underline{underlined}.}
\begin{tabular}{clccccc}
\toprule
{Datasets}\hspace{3pt}  &  {Method} & {PSNR$\uparrow$} & {SSIM$\uparrow$} & {LPIPS$\downarrow$} & {FID$\downarrow$} & {FVD$\downarrow$}  \\ %
\midrule 
\multirow{5}{*}{{RealEstate10K}} \hspace{3pt}

& \textbf{See3D}~\cite{See3D} & 14.60 & 0.5307 & 0.4402 & 38.25 & 378.4 \\ %
& \textbf{ViewCrafter}~\cite{Viewcrafter} & 14.37 & 0.4854 & 0.4670 & 32.35 & 445.1\\ %
& \textbf{FlexWorld}~\cite{FlexWorld} & 14.28 & 0.5223 & 0.4418 & \textbf{30.56} & \textbf{270.4}\\ %
& \textbf{Hunyuan-Voyager}~\cite{Voyager} & \underline{14.85} & \underline{0.5430} & \underline{0.4357} & 52.32 & 569.6 \\
& \textbf{\nameofmethod{~(ours)}} & \textbf{17.28} & \textbf{0.6193} & \textbf{0.3297} & \underline{31.00} & \underline{311.7}\\ %
\midrule

\multirow{5}{*}{{Tanks and Temples}} \hspace{3pt}

& \textbf{See3D}~\cite{See3D} & \underline{13.00} & \underline{0.3977} & 0.5400 & 53.36 & 571.9\\%
& \textbf{ViewCrafter}~\cite{Viewcrafter} & 12.53 & 0.3651 & 0.5558 & 41.33 &  716.3\\ %
& \textbf{FlexWorld}~\cite{FlexWorld} & 12.99 & 0.3938 & \underline{0.5298} & \textbf{38.69} & \textbf{422.6}\\ %
& \textbf{Hunyuan-Voyager}~\cite{Voyager} & 12.60 & 0.3855 & 0.5769 & 68.26 & 969.3 \\
& \textbf{\nameofmethod{~(ours)}}& \textbf{14.99} & \textbf{0.4625} & \textbf{0.4556} & \underline{39.39} & \underline{507.9}\\ %
\bottomrule 
\end{tabular}
\label{tab:performance_video}
\end{table*}

\begin{table*}[!t]
\setlength{\abovecaptionskip}{2pt}
\setlength\tabcolsep{6pt}
\scriptsize
\centering
\caption{
Quantitative comparison of our \nameofmethod{} with recent state-of-the-art methods on \textbf{3D scene generation}. The best performances are in \textbf{bold} and the second performances are \underline{underlined}.
}
\begin{tabular}{lcccccc}
\toprule
\multirow{2}{*}{\raisebox{-0.5ex}{Method}} & \multicolumn{3}{c}{RealEstate10K} & \multicolumn{3}{c}{Tanks and Temples} \\
\cmidrule(lr){2-4} \cmidrule(lr){5-7}
 & {PSNR$\uparrow$} & {SSIM$\uparrow$} & {LPIPS$\downarrow$} & {PSNR$\uparrow$} & {SSIM$\uparrow$} & {LPIPS$\downarrow$} \\
\midrule 
%
\textbf{See3D}~\cite{See3D} & \underline{14.67} & \underline{0.5315} & \underline{0.4413} & \underline{13.14} & \underline{0.3979} & \underline{0.5420} \\
\textbf{ViewCrafter}~\cite{Viewcrafter} & 13.34 & 0.4847 & 0.4754 & 12.02 & 0.3649 & 0.5717 \\
\textbf{FlexWorld}~\cite{FlexWorld} & 13.71 & 0.4775 & 0.5174 & 12.46 & 0.3543 & 0.5785 \\
\textbf{Hunyuan-Voyager}~\cite{Voyager} & 13.52 & 0.4780 & 0.5189 & 12.27 & 0.3528 & 0.5950 \\
\textbf{\nameofmethod{ (ours)}}& \textbf{16.64} & \textbf{0.5970} & \textbf{0.4284} & \textbf{15.00} & \textbf{0.4642} & \textbf{0.5058} \\

\bottomrule 
\end{tabular}
\label{tab:performance_scene}
\end{table*}

\myPara{Testing and evaluation.} We use the RealEstate10K (RE10K)~\cite{Re10K} and the Tanks and Temples (Tanks)~\cite{Tanks} as our test datasets, following the same construction method as FlexWorld. We randomly select 100 video clips from RE10K and 100 video clips from Tanks, each of which consists of 49 frames.
We adopt PSNR, SSIM~\cite{ssim}, LPIPS~\cite{LPIPS}, FID~\cite{FID}, and FVD~\cite{fvd} as our evaluation metrics.

\begin{figure*}[!t]
    \centering
    \setlength{\abovecaptionskip}{2pt}
    \begin{overpic}[width=\linewidth]{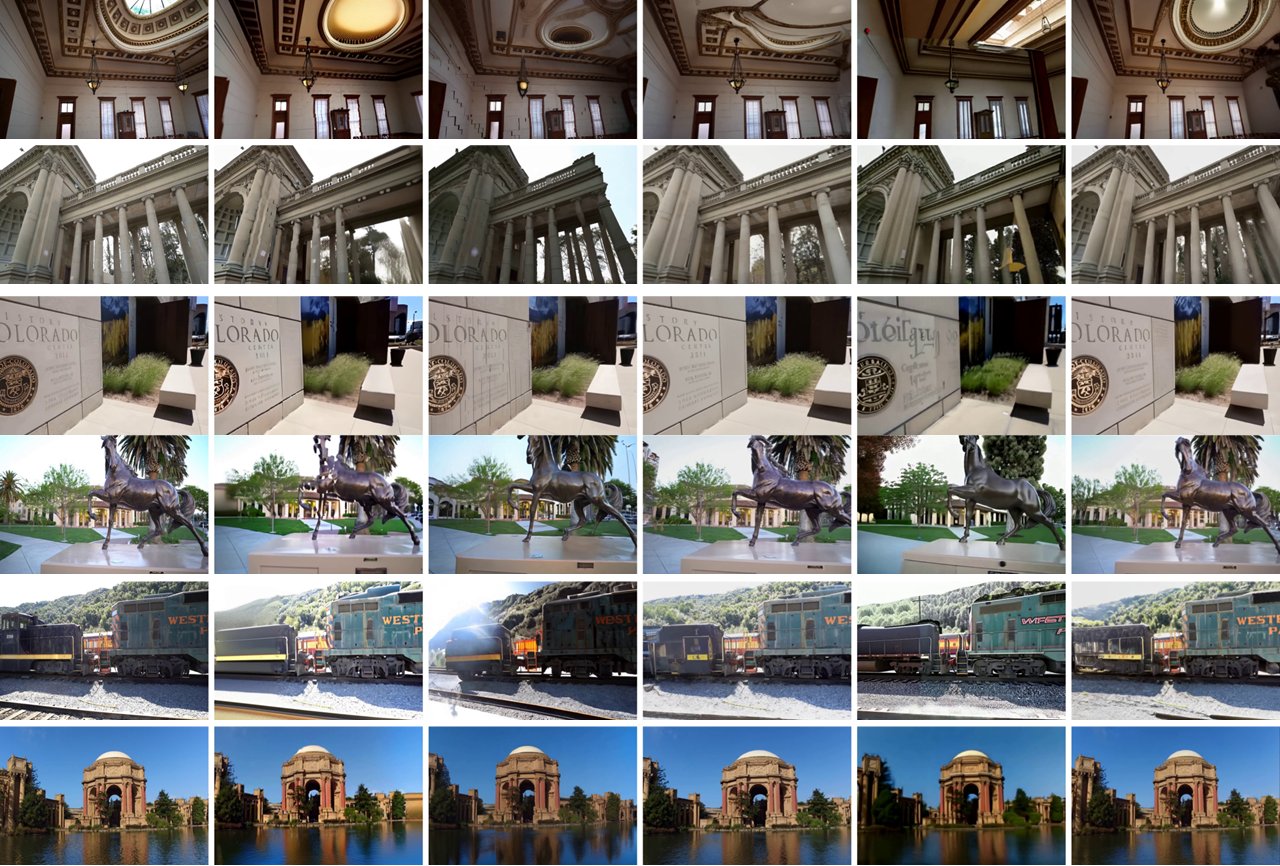}
        %
        \put(-0.2,-2){\scriptsize\textcolor{black}{(a)Ground Truth}}
        \put(20.3,-2){\scriptsize\textcolor{black}{(b)See3d}}
        \put(34,-2){\scriptsize\textcolor{black}{(c)ViewCrafter}}
        \put(51.5,-2){\scriptsize\textcolor{black}{(d)FlexWorld}}
        \put(68.9,-2){\scriptsize\textcolor{black}{(e)Voyager}}
        \put(82.1,-2){\scriptsize\textcolor{black}{(f)GeoWorld(ours)}}
    \end{overpic}
    \vspace{0pt}
    \caption{Qualitative comparison of \nameofmethod{} with state-of-the-art methods on \textbf{novel view synthesis}.}
    \label{fig:video_compares}
\end{figure*}

\begin{figure*}[!ht]
    \centering
    \begin{overpic}[width=\linewidth]{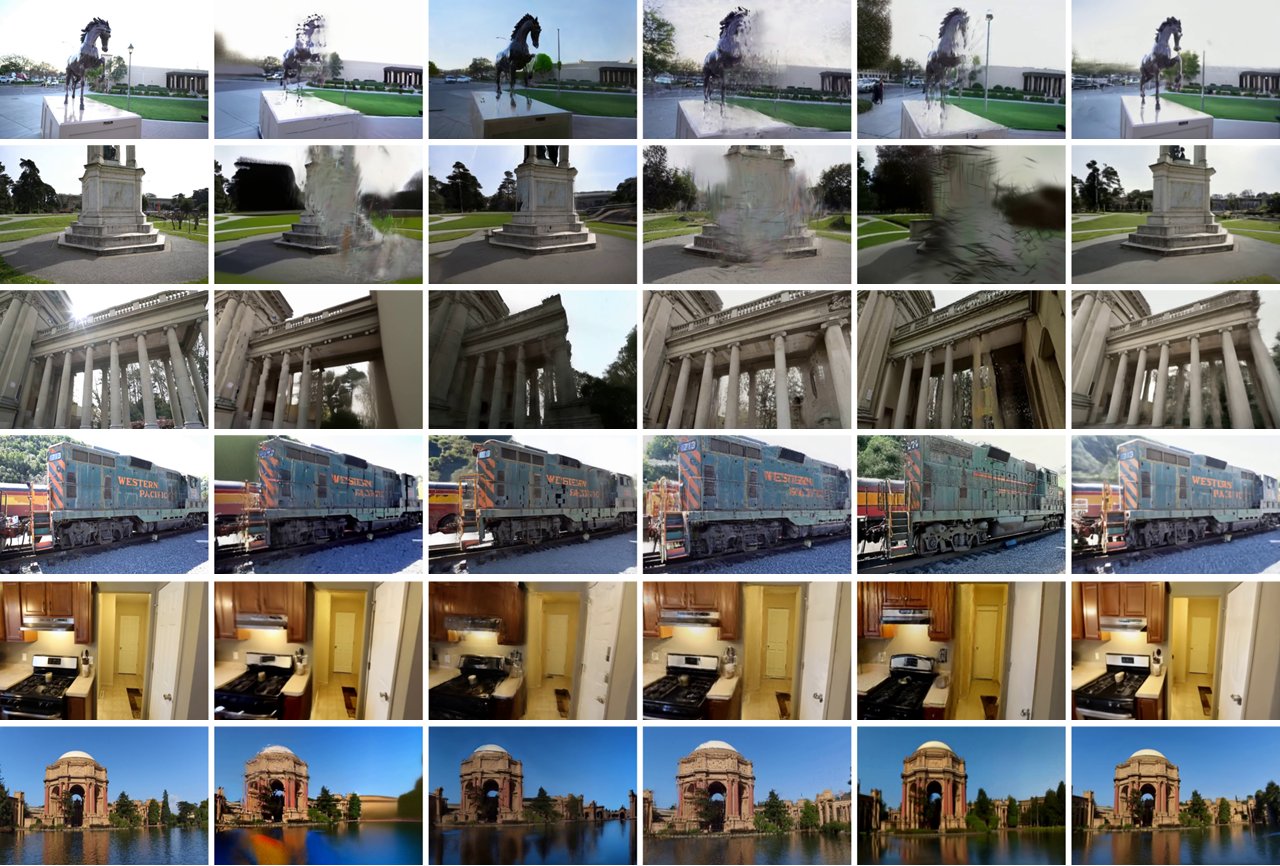}
        \put(-0.2,-2){\scriptsize\textcolor{black}{(a)Ground Truth}}
        \put(20.3,-2){\scriptsize\textcolor{black}{(b)See3d}}
        \put(34,-2){\scriptsize\textcolor{black}{(c)ViewCrafter}}
        \put(51.5,-2){\scriptsize\textcolor{black}{(d)FlexWorld}}
        \put(68.9,-2){\scriptsize\textcolor{black}{(e)Voyager}}
        \put(82.1,-2){\scriptsize\textcolor{black}{(f)GeoWorld(ours)}}
    \end{overpic}
    \caption{Qualitative comparison of our \nameofmethod{} with recent stage-of-the-art methods on \textbf{3D scene generation}.} %
    \label{fig:scene_compares}
\end{figure*}

\subsection{Comparisons with State-of-the-Art Methods}
\label{sec:compare}

\myPara{Quantitative comparisons.}
We show the quantitative comparisons between our \nameofmethod{} and recent state-of-the-art methods (See3D~\cite{See3D}, ViewCrafter~\cite{Viewcrafter}, FlexWorld~\cite{FlexWorld}, and Hunyuan-Voyager~\cite{Voyager}) in Tab.~\ref{tab:performance_video} and Tab.~\ref{tab:performance_scene}.
Tab.~\ref{tab:performance_video} reports the quantitative comparisons in novel view synthesis (i.e., the video generation process). We obtain the input of the video model following the method described in Sec.~\ref{sec:geopre}.
As shown, our \nameofmethod{} outperforms previous methods in terms of fidelity (PSNR, SSIM) and achieves competitive perceptual quality (LPIPS, FID, and FVD). In particular, the LPIPS score is lower than those of all previous methods, while the FID and FVD scores are only slightly higher than FlexWorld~\cite{FlexWorld}.

Tab.~\ref{tab:performance_scene} presents the quantitative comparisons in 3D scene generation. We reconstruct the generated videos into 3DGS and render images from corresponding camera poses to compute the evaluation metrics. %
As shown, our \nameofmethod{} surpasses previous methods across all metrics, including fidelity (PSNR, SSIM) and perceptual quality (LPIPS).
All these results demonstrate the effectiveness of our method.

\myPara{Qualitative comparisons.}
Fig.~\ref{fig:video_compares} and Fig.~\ref{fig:scene_compares} show the qualitative comparison results.
Fig.~\ref{fig:video_compares} presents the qualitative comparisons in novel view synthesis (i.e., the video generation process). As shown, the videos generated by our \nameofmethod{} contain fewer artifacts, richer details, and exhibit stable camera control.
Fig.~\ref{fig:scene_compares} shows the qualitative comparisons in 3D scene generation. As illustrated, our \nameofmethod{} can maintain high-quality geometric structures, while producing fewer artifacts compared to previous methods.

\begin{figure*}[!t]
    \centering
    \setlength{\abovecaptionskip}{2pt}
    \begin{overpic}[width=\linewidth]{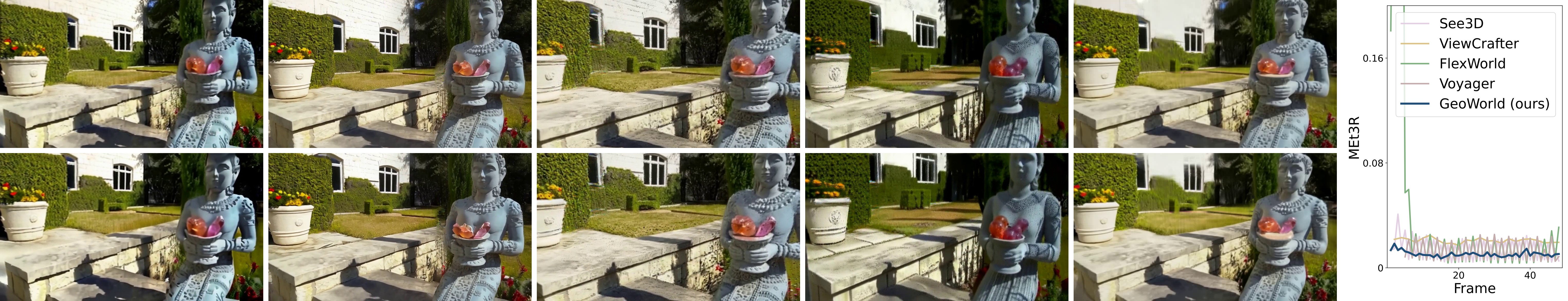}
    \end{overpic}

    \vspace{5pt}
    
    \begin{overpic}[width=\linewidth]{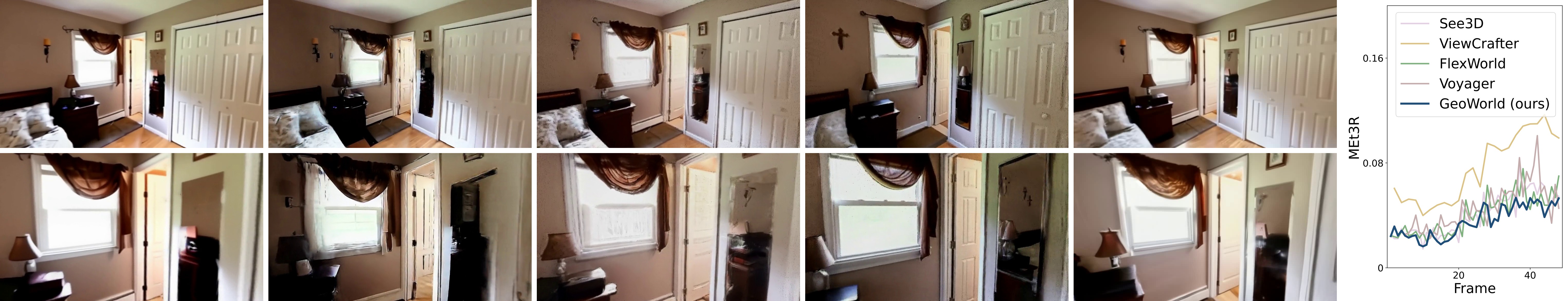}
        \put(3.9,-2){\scriptsize\textcolor{black}{(a)See3d}}
        \put(17.8,-2){\scriptsize\textcolor{black}{(b)ViewCrafter}}
        \put(35.4,-2){\scriptsize\textcolor{black}{(c)FlexWorld}}
        \put(53.5,-2){\scriptsize\textcolor{black}{(d)Voyager}}
        \put(66.3,-2){\scriptsize\textcolor{black}{(e)GeoWorld(ours)}}
        \put(85.0,-2){\scriptsize\textcolor{black}{(f)MEt3R Scores}}
    \end{overpic}
    \vspace{-3pt}
    \caption{MEt3R results~\cite{met3r}. Lower scores indicate better multi-view consistency.}
    \label{fig:3deval_met3r}
\end{figure*}

\begin{figure*}[!ht]
    \centering
    \setlength{\abovecaptionskip}{2pt}
    \begin{overpic}[width=\linewidth]{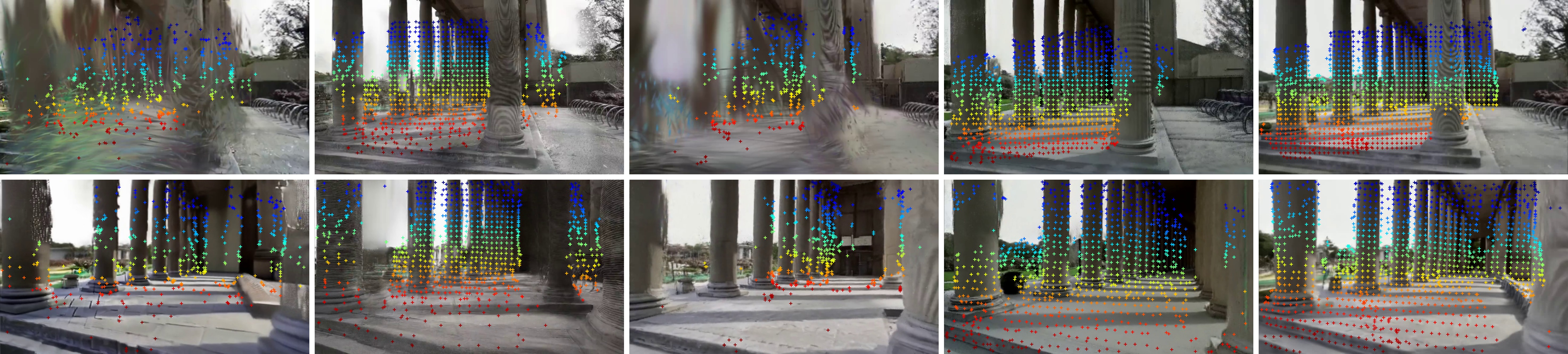}
        \put(3.8,-2){\scriptsize\textcolor{black}{Matching: 481}}
        \put(23.8,-2){\scriptsize\textcolor{black}{Matching: 890}}
        \put(43.5,-2){\scriptsize\textcolor{black}{Matching: 434}}
        \put(63.3,-2){\scriptsize\textcolor{black}{Matching: 946}}
        \put(82.2,-2){\scriptsize\textcolor{black}{Matching: \textbf{1250}}}

        \put(3.7,-4.5){\scriptsize\textcolor{black}{(a) See3d}~\cite{See3D}}
        \put(20.8,-4.5){\scriptsize\textcolor{black}{(b) ViewCrafter}~\cite{Viewcrafter}}
        \put(42.2,-4.5){\scriptsize\textcolor{black}{(c) FlexWorld}~\cite{FlexWorld}}
        \put(62.2,-4.5){\scriptsize\textcolor{black}{(d) Voyager}~\cite{Voyager}}
        \put(80.0,-4.5){\scriptsize\textcolor{black}{(e) GeoWorld (ours)}}
    \end{overpic}
    \vspace{5pt}
    \caption{Image matching results~\cite{mast3r}. `Matching' denotes the number of matched points between two views. A higher count of matches indicates superior multi-view consistency of the views.}
    \label{fig:3deval_mast3r}
\end{figure*}

\subsection{3D Evaluation}
\label{sec:3d_eval}
To further evaluate the capabilities of the image-to-3D scene model, we conduct multi-view consistency and image matching tests using MEt3R~\cite{met3r} and MASt3R~\cite{mast3r}. We reconstruct the generated videos into 3DGS and render images from corresponding camera poses to serve as the inputs for testing.

\myPara{Multi-view consistency.} 
Fig.~\ref{fig:3deval_met3r} shows the MEt3R~\cite{met3r} results. We calculate the MEt3R score for each adjacent frame in a scene (49 frames), with lower scores indicating better multi-view consistency. As shown, our GeoWorld has the lowest MEt3R scores.
Furthermore, since the videos used in the test are rendered by 3DGS, the lower MET3R score also indicates the stability of the 3DGS structure and that 3DGS contains fewer geometric conflict areas.

\myPara{Image matching.}
Fig.~\ref{fig:3deval_mast3r} shows the image matching results. We input two frames from one scene, then use MASt3R~\cite{mast3r} for image matching test and visualize the matched points. As seen, GeoWorld achieves the best matching results, and the output frames adhere to the geometric logic of 3D space at the pixel level.
These test results demonstrate the superiority of our GeoWorld and validate the rationale behind our use of geometric conditions.

\begin{table}[!t]
    \setlength\tabcolsep{1.8pt}
    \scriptsize
    \setlength{\abovecaptionskip}{2pt}
    \renewcommand{\arraystretch}{1.1}
    \centering
    \caption{Cost analysis. Our GeoWorld has lower numbers of parameters and compute cost among video diffusion model–based methods (ViewCrafter~\cite{Viewcrafter}, FlexWorld~\cite{FlexWorld}, and Hunyuan-Voyager~\cite{Voyager}). See3D~\cite{See3D} has lower GPU memory cost since it is based on a 2D diffusion model.}
        \begin{tabular}{lcccccc}
    
        \toprule
        {Model} & {PSNR(RE10K)} & {Params.} & {Train. Time} & {Infer. Size} &{Infer. Time}  &{Infer. Mem.} \\
        \hline
        See3d~\cite{See3D} &14.60 & 1.6B & 25days & 512*512*25 & 73s & 9G \\
        ViewCrafter~\cite{Viewcrafter} &14.37 & 2.6B & - & 576*1024*25 &120s & 24G \\
        FlexWorld~\cite{FlexWorld} &14.28 & 5.0B & 7days&576*1024*49 & 201s &28G \\
        Hunyuan-Voyager~\cite{Voyager}&14.85 & 12.8B & -&512*768*49 & 1110s & 48G \\
        GeoWorld (ours) & \textbf{17.28}& 3.9B & 4days&576*1024*49  & 148s & 31G
        \\
        \bottomrule
    \end{tabular}
   \label{tab:ca}
\end{table}

\subsection{Cost Analysis}
\label{sec:ca}
We report parameters (the sum of the two-stage models for GeoWorld), training time (train on 8 NVIDIA A100 80\,GB GPUs, See3D~\cite{See3D} uses 114 NVIDIA A100 40\,GB GPUs), inference time, and inference GPU memory cost in the Tab.~\ref{tab:ca} (inference on 1 NVIDIA A100 80\,GB GPU). Train time and inference time of GeoWorld represent the total duration of our two-stage pipeline. As shown, our method has lower numbers of parameters and compute cost among video diffusion model–based methods (ViewCrafter~\cite{Viewcrafter}, FlexWorld~\cite{FlexWorld}, and Hunyuan-Voyager~\cite{Voyager}), with an acceptable increase in GPU memory used to load the VGGT model during inference (See3D~\cite{See3D} has lower GPU memory cost since it is based on a 2D diffusion model).

\subsection{Ablation Study}
\label{sec:ab}
\begin{figure}[!tp]
    \centering
    \scriptsize
    \begin{overpic}[width=\linewidth]{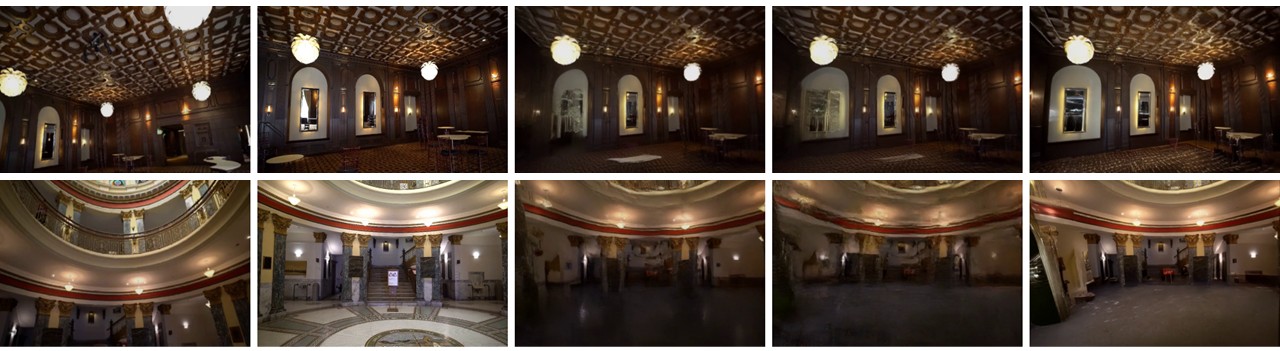}
    \put(6.5,-2){\textcolor{black}{(a)Input}}
    \put(22,-2){\textcolor{black}{(b)Ground Truth}}
    \put(41,-2){\textcolor{black}{(c)w/o geo. cond.}}
    \put(63.5,-2){\textcolor{black}{(d)depth only}}
    \put(83.5,-2){\textcolor{black}{(e)GeoWorld}}
    \end{overpic}
    \vspace{-5pt}
    \caption{Gains from the geometric conditions. As shown, providing no geometric conditions or providing simple geometric conditions for the video model are both worse than GeoWorld's method of providing high-quality geometric conditions.}
    \label{fig:gains}
\end{figure}

\myPara{Direct analysis of gains from the geometric conditions.} We perform two experiments: (i) The removal of geometric conditions throughout the entire process(Fig.~\ref{fig:gains}(c)). The purpose of (i) is to provide a straightforward `base model' to demonstrate the gains from our method. (ii) The model conditioned on depth maps(Fig.~\ref{fig:gains}(d)). The purpose of (ii) is to demonstrate that the embedding of geometric features is superior to simple geometric conditions. As shown in Fig.~\ref{fig:gains}, GeoWorld could provide clearer geometry and sharper visual content. This indicates that providing no geometric conditions or providing simple geometric conditions for the video model are both worse than GeoWorld's method of providing high-quality geometric conditions.

\begin{figure}[!ht]
    \centering
    \begin{minipage}{0.48\linewidth}
    \vspace{-9pt}
    \begin{overpic}[width=\linewidth]{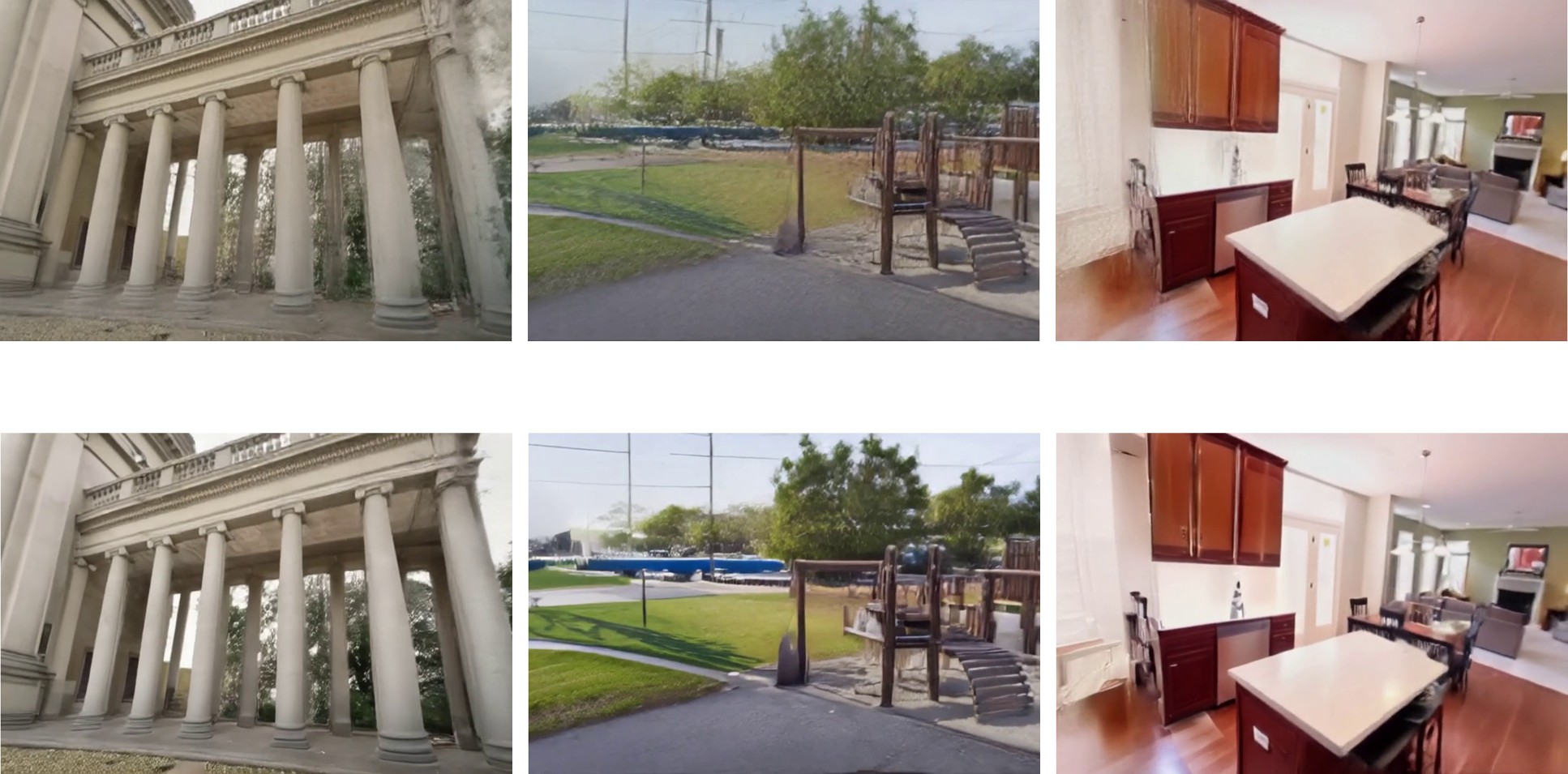}
        \put(14,23.5){\scriptsize\textcolor{black}{(a) Before geo-constrained model}}
        \put(14.5,-4.5){\scriptsize\textcolor{black}{(b) After geo-constrained model}}
    \end{overpic}
    \vspace{0pt}
    \caption{Visual comparisons of the geometry constraints.}
    \label{fig:ab_geoconstraints}
    \end{minipage}
    \hfill
    \begin{minipage}{0.48\linewidth}
    \begin{overpic}[width=\linewidth]{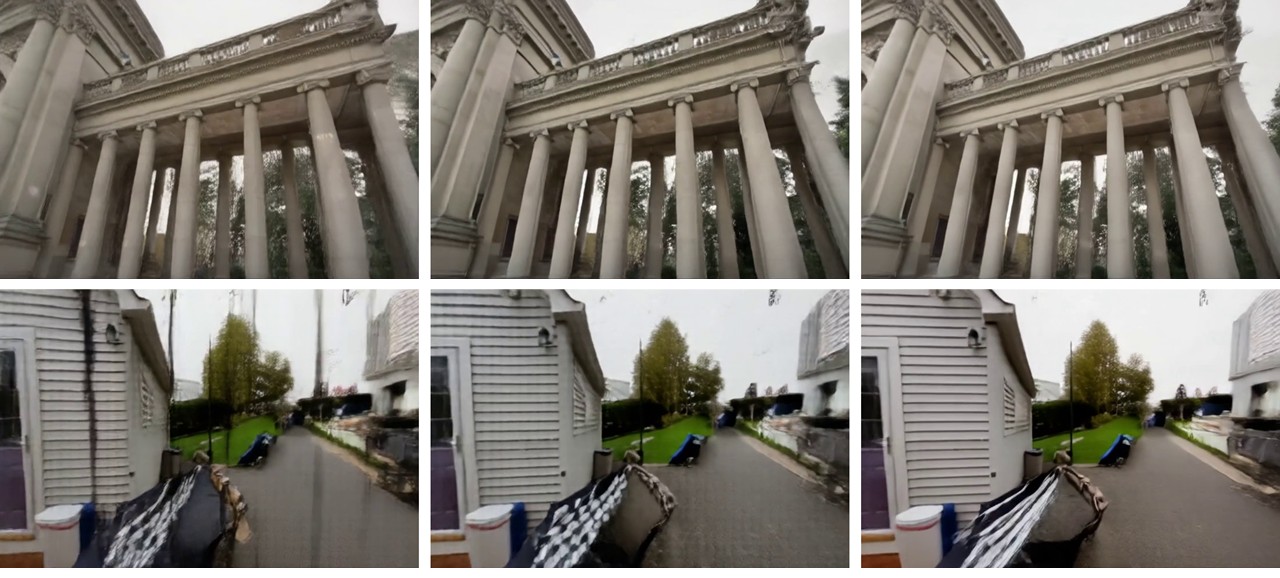}
        \put(0,-4.5){\scriptsize\textcolor{black}{(a)Emb. directly}}
        \put(40.2,-4.5){\scriptsize\textcolor{black}{(b)+GAL}}
        \put(71.8,-4.5){\scriptsize\textcolor{black}{(c)+GAM}}
    \end{overpic}
    \vspace{-5pt}
    \caption{Visual comparisons of the design of the geometry-constrained diffusion model. `GAL': geometric alignment loss. `GAM': geometry adaptation module. }
    \label{fig:ab_modeldesign}
    \end{minipage}
\end{figure}

\myPara{Effectiveness of the geometry-constrained diffusion model.} 
We then evaluate the effectiveness of geometry-constrained diffusion model by comparing the quality of the condition views produced by the geometric condition generation procedure with predicted views produced by the geometry-constrained diffusion model. Fig.~\ref{fig:ab_geoconstraints} shows the visual comparisons. It can be clearly observed that the refined videos contain almost no blurry regions, the generated objects are sharper, more detailed, and exhibit clearer geometric structures, demonstrating the effectiveness of geometric constraints.

\begin{table}[!t]
    \setlength\tabcolsep{10pt}
    \scriptsize
    \setlength{\abovecaptionskip}{2pt}
    \centering
        \caption{
      Design of geometry-constrained diffusion model.}
        \begin{tabular}{lcc}
    
        \toprule
        {Model} & {PSNR$\uparrow$} & {LPIPS$\downarrow$} \\
        \midrule
        
         Embed directly & 16.77 {\scriptsize\textcolor{gray}{(+0.00)}} & 0.3381 {\scriptsize\textcolor{gray}{(+0.0000)}}\\
         + Geometric Alignment Loss & 16.96 {\scriptsize\textcolor{gray}{(+0.19)}} & 0.3284 {\scriptsize\textcolor{gray}{(-0.0097)}} \\

         \midrule
        \textit{+ Geometry Adaptation Module} & & \\ [-1pt]  %
        \cmidrule(lr){1-1}
        + Resize  & 17.17 {\scriptsize\textcolor{gray}{(+0.40)}} & 0.3286 {\scriptsize\textcolor{gray}{(-0.0095)}} \\
        + Global Weighting & 17.28 {\scriptsize\textcolor{gray}{(+0.51)}} & 0.3292 {\scriptsize\textcolor{gray}{(-0.0089)}} \\
        \bottomrule
    \end{tabular}
    \label{tab:ab_modeldesign}
\end{table}

\myPara{Design of the geometry-constrained diffusion model.}
We then verify the rationality of our model design. The design of the geometry-constrained diffusion model consists of two main components: the geometric alignment loss described in Sec.~\ref{sec:geoloss} and the geometry adaptation module described in Sec.~\ref{sec:geoadapt}. The quantitative comparison results on the RE10K~\cite{Re10K} test set are presented in Tab.~\ref{tab:ab_modeldesign}. `Embed directly' refers to directly embedding the geometry features into the model via cross-attention. As shown, each component of our model contributes to improvements in the PSNR metric. In particular, introducing the geometric alignment loss leads to an improvement in the LPIPS metric, indicating enhanced perceptual quality, which is consistent with our objective of incorporating real-world geometric information. We also provide visual comparisons in Fig.~\ref{fig:ab_modeldesign}.
It can be observed that after introducing the geometric alignment loss, the distorted geometry and artifacts in the outputs are notably reduced, while further incorporating the geometry adaptation module leads to clearer geometric structures, reduces grid-like artifacts, and produces cleaner visual results.

\begin{figure}[t]
    \centering
    \small
    \begin{minipage}{0.5\linewidth}
        \begin{overpic}[width=\linewidth]{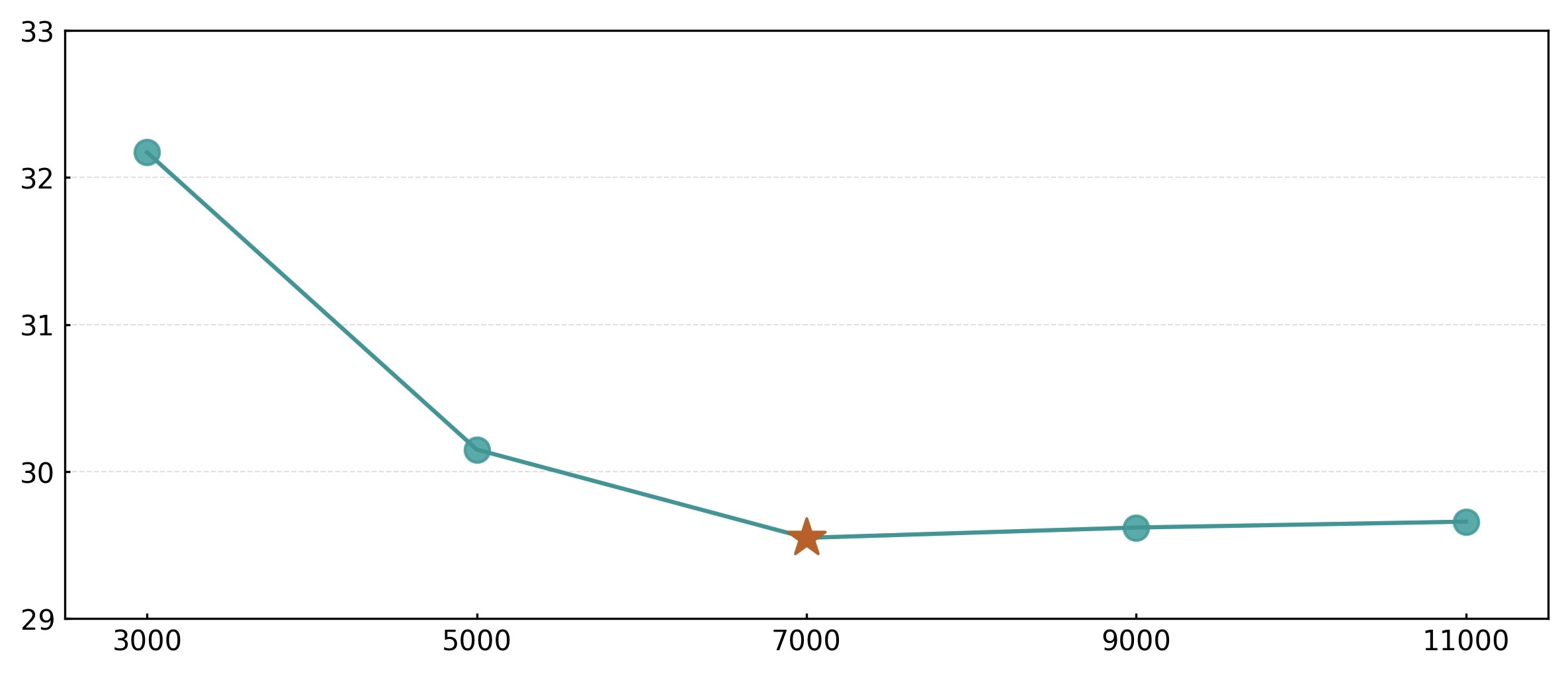}
        \put(4,42){\scriptsize\textcolor{black}{\textit{FID}}}
        \put(11,32){\scriptsize\textcolor{black}{(3000, 32.17)}}
        \put(40,11){\scriptsize\textcolor{black}{(7000, 29.55)}}
        \put(38,15){\scriptsize\textcolor{black}{[for GeoWorld]}}
        \put(75,11){\scriptsize\textcolor{black}{(11000, 29.66)}}
        \put(74,-4.5){\scriptsize\textcolor{black}{\textit{Training Step}}}
    \end{overpic}
    \caption{Quantitative comparisons of the geometric condition generation procedure.}
    \label{fig:ab_geopre}
    \end{minipage}
    \quad
    \begin{minipage}{0.45\linewidth}
    \begin{overpic}[width=\linewidth]{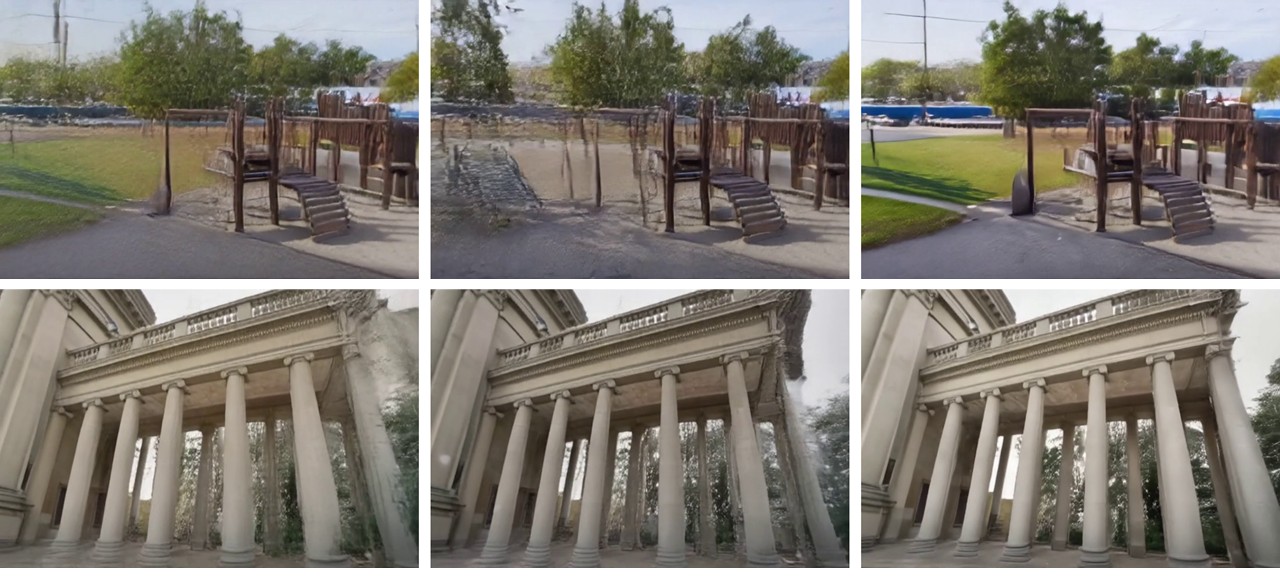}
        \put(2,-4.5){\scriptsize\textcolor{black}{(a)7000 Steps}}
        \put(34,-4.5){\scriptsize\textcolor{black}{(b)11000 Steps}}
        \put(69,-4.5){\scriptsize\textcolor{black}{(c)}}
        \put(74.8,-3.5){\scriptsize\textcolor{black}{ After geo.}}
        \put(74.8,-7){\scriptsize\textcolor{black}{constraint}}
    \end{overpic}
    \vspace{-9pt}
    \caption{Visual comparisons of the geometric condition generation procedure.}
    \label{fig:ab_geopre_visual}
    \end{minipage}

\end{figure}



\myPara{Discussion about the geometric condition generation procedure.}
The geometric condition generation procedure in our overall pipeline requires additional discussion, mainly concerning the performance of the video model fine-tuned during this process. If the fine-tuned video model is undertrained, it may fail to provide sufficient geometric information and could even introduce incorrect guidance to the geometry-constrained model. If the fine-tuned video model is trained for a sufficiently long time and can already generate highly detailed videos, the subsequent refinement process would become unnecessary. The results presented in Fig.~\ref{fig:ab_geopre} and Fig.~\ref{fig:ab_geopre_visual} might address these concerns.
As shown in Fig.~\ref{fig:ab_geopre}, the model trained for 7000 steps achieves the lowest FID, and further increasing the training steps does not lead to additional improvements. The visual comparisons in Fig.~\ref{fig:ab_geopre_visual} show that the fine-tuned video model (7000 steps) in the geometric condition generation procedure still suffers from geometric distortions and blurry content even after an additional 4000 training steps (11000 steps). In contrast, the geometry-constrained model (with an additional 2000 steps) produces videos with clearer geometry and sharper visual content.

\begin{figure}[!t]
    \centering
    \begin{minipage}{0.45\linewidth}
    
    \begin{overpic}[width=\linewidth]{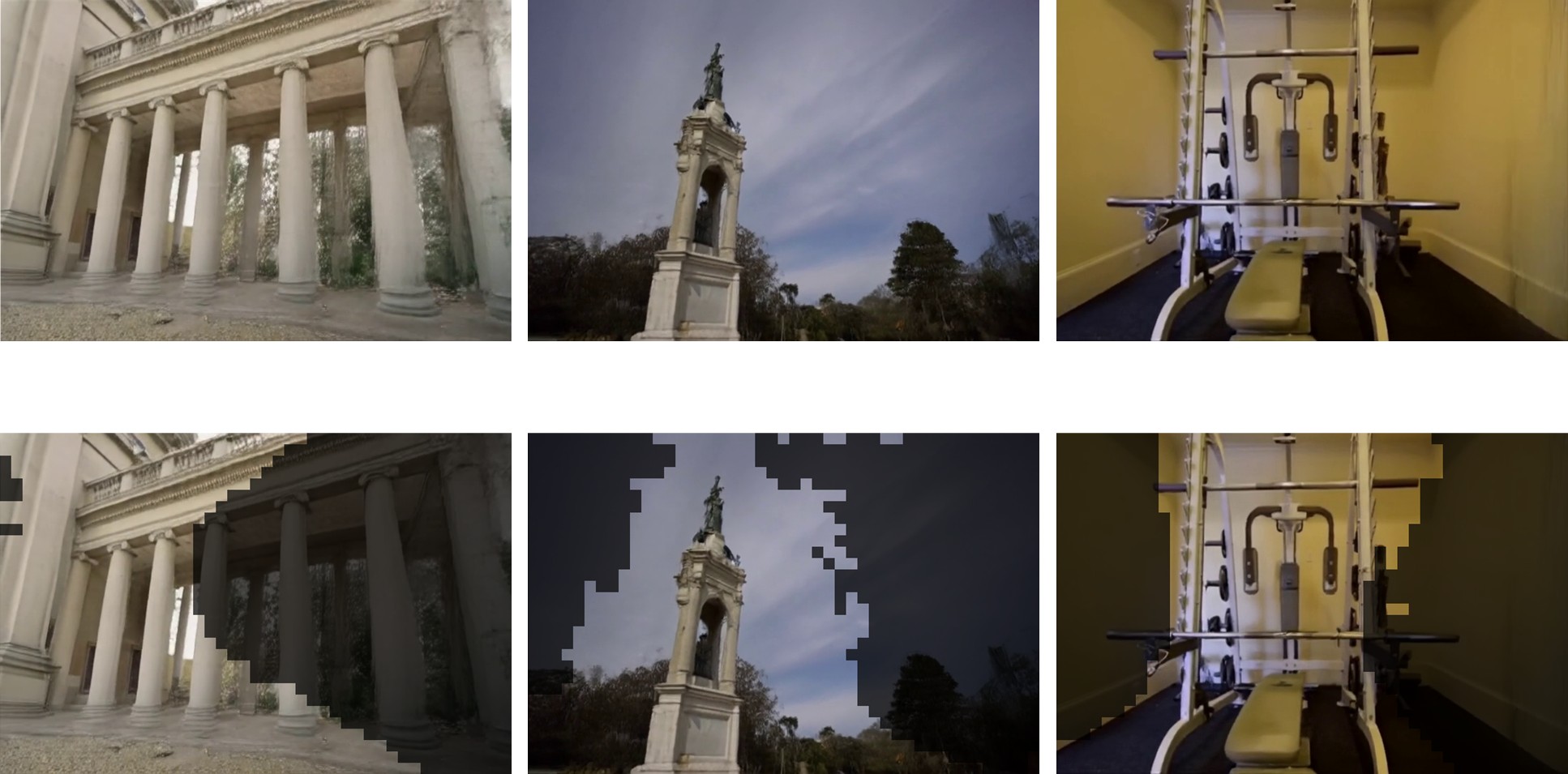}
        \put(28,23.3){\scriptsize\textcolor{black}{(a) Condition Views}}
        \put(11.5,-4.5){\scriptsize\textcolor{black}{(b) Visualization of low-weight tokens}}
    \end{overpic}
    \vspace{-10pt}
    \caption{Visualization of the global weighting process. The darker regions in the second row visualize the low-weight tokens identified by the process.}
    \label{fig:ab_globalweighting}
    \end{minipage}
    \quad
    \begin{minipage}{0.5\linewidth}
    \begin{overpic}[width=\linewidth]{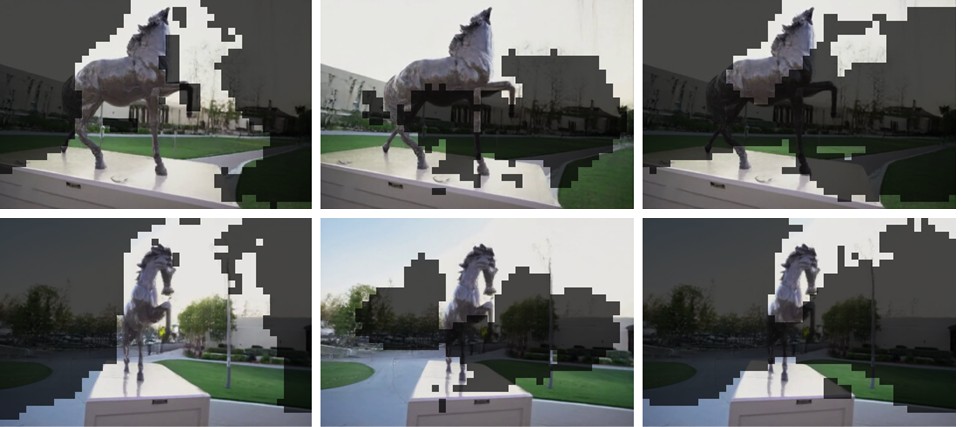}
        \put(6.8,-4.5){\scriptsize\textcolor{black}{(a) 50\%}}
        \put(39.8,-4.5){\scriptsize\textcolor{black}{(b) 30\%}}
        \put(72.8,-4.5){\scriptsize\textcolor{black}{(c) 70\%}}
    \end{overpic}
    \vspace{-8pt}
    \caption{Visual comparisons of different discard ratios in the global weighting process. The darker regions visualize the discarded tokens.}
    \label{fig:ab_sup_gwp}
    \end{minipage}
    
    
\end{figure}

\myPara{Analysis of global weighting process.}
We analyze the global weighting process proposed in Sec.~\ref{sec:geoadapt}. Fig.~\ref{fig:ab_globalweighting} presents the visualization results. The first row shows the condition views provided by the geometric condition generation procedure, while the darker regions in the second row visualize the bottom 50\% low-weight tokens identified by the global weighting process. It can be observed that our global weighting process effectively filters out low-quality tokens. As illustrated in the first column, this process is able to detect blurry regions within the condition views. In the second and third columns, when the conditional views contain few blurry regions, the process tends to retain regions containing richer geometric structures (e.g., the sky and plain-colored walls contain limited geometric information). 

\begin{table}[!t]
    \setlength\tabcolsep{10pt}
    \scriptsize
    \setlength{\abovecaptionskip}{2pt}
    \centering
        \caption{
      Quantitative results of different discard ratios in the global weighting process.}
        \begin{tabular}{lcc}
        \toprule
        {Discard ratio} & {PSNR$\uparrow$} & {SSIM$\uparrow$} \\
        \midrule
        50\% (For GeoWorld) & \textbf{17.28} {\scriptsize\textcolor{gray}{(+0.00)}} & \textbf{0.6193} {\scriptsize\textcolor{gray}{(+0.0000)}} \\
         30\% & 17.17 {\scriptsize\textcolor{gray}{(-0.11)}} & 0.6173 {\scriptsize\textcolor{gray}{(-0.0020)}}\\
         70\% & 17.22 {\scriptsize\textcolor{gray}{(-0.06)}} & 0.6188 {\scriptsize\textcolor{gray}{(-0.0005)}} \\
         \bottomrule
    \end{tabular}
    \label{tab:ab_discardratio}
\end{table}

\myPara{Ablation on the discard ratio in the global weighting process.} Fig.~\ref{fig:ab_sup_gwp} presents the visualization of the discarded tokens. As shown, when the discard ratio is set to 30\%, some low-quality regions(e.g., blurry content or regions that contain limited geometric information) are still retained. When the ratio increases to 70\%, although most low-quality regions are effectively removed, a considerable portion of the geometrically informative main regions is also discarded. A discard ratio of 50\% strikes a favorable balance between the two.
Furthermore, Tab.~\ref{tab:ab_discardratio} presents the quantitative results of different discard ratios. As can be seen, a discard ratio of 50\% achieves the best PSNR and SSIM performance.

\section{Conclusions}
In this paper, we propose \nameofmethod{}, a novel pipeline paradigm that can provide full-frame geometric features to the video generation process to alleviate the high generation difficulty caused by the limited input information inherent in this task. Furthermore, we explore how to leverage geometry models to obtain full-frame features and assist the video generation process.
Extensive experiments demonstrate the effectiveness of our method, which outperforms previous methods qualitatively while achieving superior fidelity and competitive perceptual quality quantitatively.
We expect that our \nameofmethod{} can inspire future research and provide a new perspective for designing image-to-3D scene generation models.

\myPara{Acknowledgments.} This work was funded by the National Natural Science Foundation of China under 62522607, 62495061, and 625B2093, and the Fundamental Research Funds for
the Central Universities (Nankai University).

%
%
\bibliographystyle{splncs04}
\bibliography{main}

@String(ICLR  = {Int. Conf. Learn. Represent.})

@String(TOG   = {ACM Trans. Graph.})

@String(ICLR  = {ICLR})

@String(TOG   = {ACM TOG})

@article{DDPM,
  title={Denoising diffusion probabilistic models},
  author={Ho, Jonathan and Jain, Ajay and Abbeel, Pieter},
  journal={Advances in neural information processing systems},
  volume={33},
  pages={6840--6851},
  year={2020}
}

@inproceedings{LDM,
  title={High-resolution image synthesis with latent diffusion models},
  author={Rombach, Robin and Blattmann, Andreas and Lorenz, Dominik and Esser, Patrick and Ommer, Bj{\"o}rn},
  booktitle={Proceedings of the IEEE/CVF conference on computer vision and pattern recognition},
  pages={10684--10695},
  year={2022}
}

@inproceedings{Set-the-scene,
  title={Set-the-scene: Global-local training for generating controllable nerf scenes},
  author={Cohen-Bar, Dana and Richardson, Elad and Metzer, Gal and Giryes, Raja and Cohen-Or, Daniel},
  booktitle={Proceedings of the IEEE/CVF International Conference on Computer Vision},
  pages={2920--2929},
  year={2023}
}

@inproceedings{Dreamscene,
  title={Dreamscene: 3d gaussian-based text-to-3d scene generation via formation pattern sampling},
  author={Li, Haoran and Shi, Haolin and Zhang, Wenli and Wu, Wenjun and Liao, Yong and Wang, Lin and Lee, Lik-hang and Zhou, Peng Yuan},
  booktitle={European Conference on Computer Vision},
  pages={214--230},
  year={2024},
  organization={Springer}
}

@article{Luciddreamer,
  title={Luciddreamer: Domain-free generation of 3d gaussian splatting scenes},
  author={Chung, Jaeyoung and Lee, Suyoung and Nam, Hyeongjin and Lee, Jaerin and Lee, Kyoung Mu},
  journal={arXiv preprint arXiv:2311.13384},
  year={2023}
}

@inproceedings{Wonderworld,
  title={Wonderworld: Interactive 3d scene generation from a single image},
  author={Yu, Hong-Xing and Duan, Haoyi and Herrmann, Charles and Freeman, William T and Wu, Jiajun},
  booktitle={Proceedings of the Computer Vision and Pattern Recognition Conference},
  pages={5916--5926},
  year={2025}
}

@inproceedings{Dreamscene360,
  title={Dreamscene360: Unconstrained text-to-3d scene generation with panoramic gaussian splatting},
  author={Zhou, Shijie and Fan, Zhiwen and Xu, Dejia and Chang, Haoran and Chari, Pradyumna and Bharadwaj, Tejas and You, Suya and Wang, Zhangyang and Kadambi, Achuta},
  booktitle={European Conference on Computer Vision},
  pages={324--342},
  year={2024},
  organization={Springer}
}

@inproceedings{Text2room,
  title={Text2room: Extracting textured 3d meshes from 2d text-to-image models},
  author={H{\"o}llein, Lukas and Cao, Ang and Owens, Andrew and Johnson, Justin and Nie{\ss}ner, Matthias},
  booktitle={Proceedings of the IEEE/CVF International Conference on Computer Vision},
  pages={7909--7920},
  year={2023}
}

@inproceedings{Cogvideox,
  title={Cogvideox: Text-to-video diffusion models with an expert transformer},
  author={Yang, Zhuoyi and Teng, Jiayan and Zheng, Wendi and Ding, Ming and Huang, Shiyu and Xu, Jiazheng and Yang, Yuanming and Hong, Wenyi and Zhang, Xiaohan and Feng, Guanyu and others},
  booktitle={International Conference on Learning Representations},
  volume={2025},
  pages={83048--83077},
  year={2025}
}

@article{Cogvideo,
  title={Cogvideo: Large-scale pretraining for text-to-video generation via transformers},
  author={Hong, Wenyi and Ding, Ming and Zheng, Wendi and Liu, Xinghan and Tang, Jie},
  journal={arXiv preprint arXiv:2205.15868},
  year={2022}
}

@article{Wan,
  title={Wan: Open and advanced large-scale video generative models},
  author={Wan, Team and Wang, Ang and Ai, Baole and Wen, Bin and Mao, Chaojie and Xie, Chen-Wei and Chen, Di and Yu, Feiwu and Zhao, Haiming and Yang, Jianxiao and others},
  journal={arXiv preprint arXiv:2503.20314},
  year={2025}
}

@article{Viewcrafter,
  title={Viewcrafter: Taming video diffusion models for high-fidelity novel view synthesis},
  author={Yu, Wangbo and Xing, Jinbo and Yuan, Li and Hu, Wenbo and Li, Xiaoyu and Huang, Zhipeng and Gao, Xiangjun and Wong, Tien-Tsin and Shan, Ying and Tian, Yonghong},
  journal={arXiv preprint arXiv:2409.02048},
  year={2024}
}

@article{Dimensionx,
  title={Dimensionx: Create any 3d and 4d scenes from a single image with controllable video diffusion},
  author={Sun, Wenqiang and Chen, Shuo and Liu, Fangfu and Chen, Zilong and Duan, Yueqi and Zhang, Jun and Wang, Yikai},
  journal={arXiv preprint arXiv:2411.04928},
  year={2024}
}

@article{FlexWorld,
  title={FlexWorld: Progressively expanding 3D scenes for flexiable-view synthesis},
  author={Chen, Luxi and Zhou, Zihan and Zhao, Min and Wang, Yikai and Zhang, Ge and Huang, Wenhao and Sun, Hao and Wen, Ji-Rong and Li, Chongxuan},
  journal={arXiv preprint arXiv:2503.13265},
  year={2025}
}

@article{Voyager,
  title={Voyager: Long-range and world-consistent video diffusion for explorable 3d scene generation},
  author={Huang, Tianyu and Zheng, Wangguandong and Wang, Tengfei and Liu, Yuhao and Wang, Zhenwei and Wu, Junta and Jiang, Jie and Li, Hui and Lau, Rynson and Zuo, Wangmeng and others},
  journal={ACM Transactions on Graphics (TOG)},
  volume={44},
  number={6},
  pages={1--15},
  year={2025},
  publisher={ACM New York, NY, USA}
}

@inproceedings{Stargen,
  title={Stargen: A spatiotemporal autoregression framework with video diffusion model for scalable and controllable scene generation},
  author={Zhai, Shangjin and Ye, Zhichao and Liu, Jialin and Xie, Weijian and Hu, Jiaqi and Peng, Zhen and Xue, Hua and Chen, Danpeng and Wang, Xiaomeng and Yang, Lei and others},
  booktitle={Proceedings of the Computer Vision and Pattern Recognition Conference},
  pages={26822--26833},
  year={2025}
}

@inproceedings{VGGT,
  title={Vggt: Visual geometry grounded transformer},
  author={Wang, Jianyuan and Chen, Minghao and Karaev, Nikita and Vedaldi, Andrea and Rupprecht, Christian and Novotny, David},
  booktitle={Proceedings of the Computer Vision and Pattern Recognition Conference},
  pages={5294--5306},
  year={2025}
}

@article{3DGS,
  title={3D Gaussian splatting for real-time radiance field rendering.},
  author={Kerbl, Bernhard and Kopanas, Georgios and Leimk{\"u}hler, Thomas and Drettakis, George},
  journal={ACM Trans. Graph.},
  volume={42},
  number={4},
  pages={139--1},
  year={2023}
}

@inproceedings{Dl3dv,
  title={Dl3dv-10k: A large-scale scene dataset for deep learning-based 3d vision},
  author={Ling, Lu and Sheng, Yichen and Tu, Zhi and Zhao, Wentian and Xin, Cheng and Wan, Kun and Yu, Lantao and Guo, Qianyu and Yu, Zixun and Lu, Yawen and others},
  booktitle={Proceedings of the IEEE/CVF Conference on Computer Vision and Pattern Recognition},
  pages={22160--22169},
  year={2024}
}

@article{Re10K,
  title={Stereo magnification: Learning view synthesis using multiplane images},
  author={Zhou, Tinghui and Tucker, Richard and Flynn, John and Fyffe, Graham and Snavely, Noah},
  journal={arXiv preprint arXiv:1805.09817},
  year={2018}
}

@article{Tanks,
  title={Tanks and temples: Benchmarking large-scale scene reconstruction},
  author={Knapitsch, Arno and Park, Jaesik and Zhou, Qian-Yi and Koltun, Vladlen},
  journal={ACM Transactions on Graphics (ToG)},
  volume={36},
  number={4},
  pages={1--13},
  year={2017},
  publisher={ACM New York, NY, USA}
}

@article{FID,
  title={Gans trained by a two time-scale update rule converge to a local nash equilibrium},
  author={Heusel, Martin and Ramsauer, Hubert and Unterthiner, Thomas and Nessler, Bernhard and Hochreiter, Sepp},
  journal={Advances in neural information processing systems},
  volume={30},
  year={2017}
}

@inproceedings{fvd,
  title={FVD: A new metric for video generation},
  author={Unterthiner, Thomas and Van Steenkiste, Sjoerd and Kurach, Karol and Marinier, Rapha{\"e}l and Michalski, Marcin and Gelly, Sylvain},
  booktitle = {ICLR Workshop},
  year={2019}
}

@article{ssim,
  title={Image quality assessment: from error visibility to structural similarity},
  author={Wang, Zhou and Bovik, Alan C and Sheikh, Hamid R and Simoncelli, Eero P},
  journal={IEEE transactions on image processing},
  volume={13},
  number={4},
  pages={600--612},
  year={2004},
  publisher={IEEE}
}

@inproceedings{LPIPS,
  title={The unreasonable effectiveness of deep features as a perceptual metric},
  author={Zhang, Richard and Isola, Phillip and Efros, Alexei A and Shechtman, Eli and Wang, Oliver},
  booktitle={Proceedings of the IEEE conference on computer vision and pattern recognition},
  pages={586--595},
  year={2018}
}

@inproceedings{SEVA,
  title={Stable virtual camera: Generative view synthesis with diffusion models},
  author={Zhou, Jensen and Gao, Hang and Voleti, Vikram and Vasishta, Aaryaman and Yao, Chun-Han and Boss, Mark and Torr, Philip and Rupprecht, Christian and Jampani, Varun},
  booktitle={Proceedings of the IEEE/CVF International Conference on Computer Vision},
  pages={12405--12414},
  year={2025}
}

@inproceedings{See3D,
  title={You see it, you got it: Learning 3d creation on pose-free videos at scale},
  author={Ma, Baorui and Gao, Huachen and Deng, Haoge and Luo, Zhengxiong and Huang, Tiejun and Tang, Lulu and Wang, Xinlong},
  booktitle={Proceedings of the Computer Vision and Pattern Recognition Conference},
  pages={2016--2029},
  year={2025}
}

@article{Cameractrl,
  title={Cameractrl: Enabling camera control for text-to-video generation},
  author={He, Hao and Xu, Yinghao and Guo, Yuwei and Wetzstein, Gordon and Dai, Bo and Li, Hongsheng and Yang, Ceyuan},
  journal={arXiv preprint arXiv:2404.02101},
  year={2024}
}

@inproceedings{Wonderland,
  title={Wonderland: Navigating 3d scenes from a single image},
  author={Liang, Hanwen and Cao, Junli and Goel, Vidit and Qian, Guocheng and Korolev, Sergei and Terzopoulos, Demetri and Plataniotis, Konstantinos N and Tulyakov, Sergey and Ren, Jian},
  booktitle={Proceedings of the Computer Vision and Pattern Recognition Conference},
  pages={798--810},
  year={2025}
}

@inproceedings{Dust3r,
  title={Dust3r: Geometric 3d vision made easy},
  author={Wang, Shuzhe and Leroy, Vincent and Cabon, Yohann and Chidlovskii, Boris and Revaud, Jerome},
  booktitle={Proceedings of the IEEE/CVF Conference on Computer Vision and Pattern Recognition},
  pages={20697--20709},
  year={2024}
}

@inproceedings{mast3r,
  title={Grounding image matching in 3d with mast3r},
  author={Leroy, Vincent and Cabon, Yohann and Revaud, J{\'e}r{\^o}me},
  booktitle={European Conference on Computer Vision},
  pages={71--91},
  year={2024},
  organization={Springer}
}

@inproceedings{Fast3r,
  title={Fast3r: Towards 3d reconstruction of 1000+ images in one forward pass},
  author={Yang, Jianing and Sax, Alexander and Liang, Kevin J and Henaff, Mikael and Tang, Hao and Cao, Ang and Chai, Joyce and Meier, Franziska and Feiszli, Matt},
  booktitle={Proceedings of the Computer Vision and Pattern Recognition Conference},
  pages={21924--21935},
  year={2025}
}

@inproceedings{GenRC,
  title={GenRC: Generative 3D room completion from sparse image collections},
  author={Li, Ming-Feng and Ku, Yueh-Feng and Yen, Hong-Xuan and Liu, Chi and Liu, Yu-Lun and Chen, Albert YC and Kuo, Cheng-Hao and Sun, Min},
  booktitle={European Conference on Computer Vision},
  pages={146--163},
  year={2024},
  organization={Springer}
}

@inproceedings{Layerpano3d,
  title={Layerpano3d: Layered 3d panorama for hyper-immersive scene generation},
  author={Yang, Shuai and Tan, Jing and Zhang, Mengchen and Wu, Tong and Wetzstein, Gordon and Liu, Ziwei and Lin, Dahua},
  booktitle={Proceedings of the Special Interest Group on Computer Graphics and Interactive Techniques Conference Conference Papers},
  pages={1--10},
  year={2025}
}

@article{Director3d,
  title={Director3d: Real-world camera trajectory and 3d scene generation from text},
  author={Li, Xinyang and Lai, Zhangyu and Xu, Linning and Qu, Yansong and Cao, Liujuan and Zhang, Shengchuan and Dai, Bo and Ji, Rongrong},
  journal={Advances in neural information processing systems},
  volume={37},
  pages={75125--75151},
  year={2024}
}

@inproceedings{Gs-lrm,
  title={Gs-lrm: Large reconstruction model for 3d gaussian splatting},
  author={Zhang, Kai and Bi, Sai and Tan, Hao and Xiangli, Yuanbo and Zhao, Nanxuan and Sunkavalli, Kalyan and Xu, Zexiang},
  booktitle={European Conference on Computer Vision},
  pages={1--19},
  year={2024},
  organization={Springer}
}

@inproceedings{Flash3d,
  title={Flash3d: Feed-forward generalisable 3d scene reconstruction from a single image},
  author={Szymanowicz, Stanislaw and Insafutdinov, Eldar and Zheng, Chuanxia and Campbell, Dylan and Henriques, Joao F and Rupprecht, Christian and Vedaldi, Andrea},
  booktitle={2025 International Conference on 3D Vision (3DV)},
  pages={670--681},
  year={2025},
  organization={IEEE}
}

@inproceedings{Midi,
  title={Midi: Multi-instance diffusion for single image to 3d scene generation},
  author={Huang, Zehuan and Guo, Yuan-Chen and An, Xingqiao and Yang, Yunhan and Li, Yangguang and Zou, Zi-Xin and Liang, Ding and Liu, Xihui and Cao, Yan-Pei and Sheng, Lu},
  booktitle={Proceedings of the Computer Vision and Pattern Recognition Conference},
  pages={23646--23657},
  year={2025}
}

@article{Architect,
  title={Architect: Generating vivid and interactive 3d scenes with hierarchical 2d inpainting},
  author={Wang, Yian and Qiu, Xiaowen and Liu, Jiageng and Chen, Zhehuan and Cai, Jiting and Wang, Yufei and Wang, Tsun-Hsuan Johnson and Xian, Zhou and Gan, Chuang},
  journal={Advances in Neural Information Processing Systems},
  volume={37},
  pages={67575--67603},
  year={2024}
}

@article{Cast,
  title={Cast: Component-aligned 3d scene reconstruction from an rgb image},
  author={Yao, Kaixin and Zhang, Longwen and Yan, Xinhao and Zeng, Yan and Zhang, Qixuan and Xu, Lan and Yang, Wei and Gu, Jiayuan and Yu, Jingyi},
  journal={ACM Transactions on Graphics (TOG)},
  volume={44},
  number={4},
  pages={1--19},
  year={2025},
  publisher={ACM New York, NY, USA}
}

@article{dosovitskiy2020image,
  title={An image is worth 16x16 words: Transformers for image recognition at scale},
  author={Dosovitskiy, Alexey and Beyer, Lucas and Kolesnikov, Alexander and Weissenborn, Dirk and Zhai, Xiaohua and Unterthiner, Thomas and Dehghani, Mostafa and Minderer, Matthias and Heigold, Georg and Gelly, Sylvain and others},
  journal={arXiv preprint arXiv:2010.11929},
  year={2020}
}

@inproceedings{SENet,
  title={Squeeze-and-excitation networks},
  author={Hu, Jie and Shen, Li and Sun, Gang},
  booktitle={Proceedings of the IEEE conference on computer vision and pattern recognition},
  pages={7132--7141},
  year={2018}
}

@article{IDCNet,
  title={IDCNet: Guided Video Diffusion for Metric-Consistent RGBD Scene Generation with Precise Camera Control},
  author={Liu, Lijuan and Li, Wenfa and Zhang, Dongbo and Wang, Shuo and Jiao, Shaohui},
  journal={arXiv preprint arXiv:2508.04147},
  year={2025}
}

@inproceedings{Dist-4d,
  title={Dist-4d: Disentangled spatiotemporal diffusion with metric depth for 4d driving scene generation},
  author={Guo, Jiazhe and Ding, Yikang and Chen, Xiwu and Chen, Shuo and Li, Bohan and Zou, Yingshuang and Lyu, Xiaoyang and Tan, Feiyang and Qi, Xiaojuan and Li, Zhiheng and others},
  booktitle={Proceedings of the IEEE/CVF International Conference on Computer Vision},
  pages={27231--27241},
  year={2025}
}

@inproceedings{AR-1-to-3,
  title={AR-1-to-3: Single Image to Consistent 3D Object via Next-View Prediction},
  author={Zhang, Xuying and Zhou, Yupeng and Wang, Kai and Wang, Yikai and Li, Zhen and Jiao, Shaohui and Zhou, Daquan and Hou, Qibin and Cheng, Ming-Ming},
  booktitle={Proceedings of the IEEE/CVF International Conference on Computer Vision},
  pages={26273--26283},
  year={2025}
}

@inproceedings{Cut3r,
  title={Continuous 3d perception model with persistent state},
  author={Wang, Qianqian and Zhang, Yifei and Holynski, Aleksander and Efros, Alexei A and Kanazawa, Angjoo},
  booktitle={Proceedings of the Computer Vision and Pattern Recognition Conference},
  pages={10510--10522},
  year={2025}
}

@article{pi3,
  title={$\pi^3$: Permutation-Equivariant Visual Geometry Learning},
  author={Wang, Yifan and Zhou, Jianjun and Zhu, Haoyi and Chang, Wenzheng and Zhou, Yang and Li, Zizun and Chen, Junyi and Pang, Jiangmiao and Shen, Chunhua and He, Tong},
  journal={arXiv preprint arXiv:2507.13347},
  year={2025}
}

@inproceedings{flare,
  title={Flare: Feed-forward geometry, appearance and camera estimation from uncalibrated sparse views},
  author={Zhang, Shangzhan and Wang, Jianyuan and Xu, Yinghao and Xue, Nan and Rupprecht, Christian and Zhou, Xiaowei and Shen, Yujun and Wetzstein, Gordon},
  booktitle={Proceedings of the Computer Vision and Pattern Recognition Conference},
  pages={21936--21947},
  year={2025}
}

@article{Dreamfusion,
  title={Dreamfusion: Text-to-3d using 2d diffusion},
  author={Poole, Ben and Jain, Ajay and Barron, Jonathan T and Mildenhall, Ben},
  journal={arXiv preprint arXiv:2209.14988},
  year={2022}
}

@inproceedings{Uniscene,
  title={Uniscene: Unified occupancy-centric driving scene generation},
  author={Li, Bohan and Guo, Jiazhe and Liu, Hongsi and Zou, Yingshuang and Ding, Yikang and Chen, Xiwu and Zhu, Hu and Tan, Feiyang and Zhang, Chi and Wang, Tiancai and others},
  booktitle={Proceedings of the Computer Vision and Pattern Recognition Conference},
  pages={11971--11981},
  year={2025}
}

@inproceedings{Magicdrive,
  title={Magicdrive: Street view generation with diverse 3d geometry control},
  author={Gao, Ruiyuan and Chen, Kai and Xie, Enze and Hong, Lanqing and Li, Zhenguo and Yeung, Dit-Yan and Xu, Qiang},
  booktitle={International Conference on Learning Representations},
  volume={2024},
  pages={22841--22860},
  year={2024}
}

@inproceedings{DiffusionGS,
  title={Baking gaussian splatting into diffusion denoiser for fast and scalable single-stage image-to-3d generation and reconstruction},
  author={Cai, Yuanhao and Zhang, He and Zhang, Kai and Liang, Yixun and Ren, Mengwei and Luan, Fujun and Liu, Qing and Kim, Soo Ye and Zhang, Jianming and Zhang, Zhifei and others},
  booktitle={Proceedings of the IEEE/CVF International Conference on Computer Vision},
  pages={25062--25072},
  year={2025}
}

@article{cat3d,
  title={Cat3d: Create anything in 3d with multi-view diffusion models},
  author={Gao, Ruiqi and Holynski, Aleksander and Henzler, Philipp and Brussee, Arthur and Martin-Brualla, Ricardo and Srinivasan, Pratul and Barron, Jonathan T and Poole, Ben},
  journal={arXiv preprint arXiv:2405.10314},
  year={2024}
}

@inproceedings{SceneWiz3D,
  title={Towards text-guided 3d scene composition},
  author={Zhang, Qihang and Wang, Chaoyang and Siarohin, Aliaksandr and Zhuang, Peiye and Xu, Yinghao and Yang, Ceyuan and Lin, Dahua and Zhou, Bolei and Tulyakov, Sergey and Lee, Hsin-Ying},
  booktitle={Proceedings of the IEEE/CVF Conference on Computer Vision and Pattern Recognition},
  pages={6829--6838},
  year={2024}
}

@article{Fastscene,
  title={Fastscene: Text-driven fast 3d indoor scene generation via panoramic gaussian splatting},
  author={Ma, Yikun and Zhan, Dandan and Jin, Zhi},
  journal={arXiv preprint arXiv:2405.05768},
  year={2024}
}

@inproceedings{Wonderjourney,
  title={Wonderjourney: Going from anywhere to everywhere},
  author={Yu, Hong-Xing and Duan, Haoyi and Hur, Junhwa and Sargent, Kyle and Rubinstein, Michael and Freeman, William T and Cole, Forrester and Sun, Deqing and Snavely, Noah and Wu, Jiajun and others},
  booktitle={Proceedings of the IEEE/CVF Conference on Computer Vision and Pattern Recognition},
  pages={6658--6667},
  year={2024}
}

@inproceedings{Gen3dsr,
  title={Gen3dsr: Generalizable 3d scene reconstruction via divide and conquer from a single view},
  author={Ardelean, Andreea and {\"O}zer, Mert and Egger, Bernhard},
  booktitle={2025 International Conference on 3D Vision (3DV)},
  pages={616--626},
  year={2025},
  organization={IEEE}
}

@inproceedings{Genxd,
  title={Genxd: Generating any 3d and 4d scenes},
  author={Zhao, Yuyang and Lin, Chung-Ching and Lin, Kevin and Yan, Zhiwen and Li, Linjie and Yang, Zhengyuan and Wang, Jianfeng and Lee, Gim H and Wang, Lijuan},
  booktitle={International Conference on Learning Representations},
  volume={2025},
  pages={95479--95499},
  year={2025}
}

@inproceedings{HiScene,
  title={HiScene: creating hierarchical 3D scenes with isometric view generation},
  author={Dong, Wenqi and Yang, Bangbang and Yang, Zesong and Li, Yuan and Hu, Tao and Bao, Hujun and Ma, Yuewen and Cui, Zhaopeng},
  booktitle={Proceedings of the 33rd ACM International Conference on Multimedia},
  pages={9783--9792},
  year={2025}
}

@inproceedings{Rgbd2,
  title={Rgbd2: Generative scene synthesis via incremental view inpainting using rgbd diffusion models},
  author={Lei, Jiabao and Tang, Jiapeng and Jia, Kui},
  booktitle={Proceedings of the IEEE/CVF conference on computer vision and pattern recognition},
  pages={8422--8434},
  year={2023}
}

@inproceedings{met3r,
  title={Met3r: Measuring multi-view consistency in generated images},
  author={Asim, Mohammad and Wewer, Christopher and Wimmer, Thomas and Schiele, Bernt and Lenssen, Jan Eric},
  booktitle={Proceedings of the IEEE/CVF Conference on Computer Vision and Pattern Recognition},
  pages={6034--6044},
  year={2025}
}

@article{flashworld,
  title={FlashWorld: High-quality 3D Scene Generation within Seconds},
  author={Li, Xinyang and Wang, Tengfei and Gu, Zixiao and Zhang, Shengchuan and Guo, Chunchao and Cao, Liujuan},
  journal={arXiv preprint arXiv:2510.13678},
  year={2025}
}

@article{Matrixgame,
  title={Matrix-game 2.0: An open-source real-time and streaming interactive world model},
  author={He, Xianglong and Peng, Chunli and Liu, Zexiang and Wang, Boyang and Zhang, Yifan and Cui, Qi and Kang, Fei and Jiang, Biao and An, Mengyin and Ren, Yangyang and others},
  journal={arXiv preprint arXiv:2508.13009},
  year={2025}
}

@article{yan,
  title={Yan: Foundational interactive video generation},
  author={Ye, Deheng and Zhou, Fangyun and Lv, Jiacheng and Ma, Jianqi and Zhang, Jun and Lv, Junyan and Li, Junyou and Deng, Minwen and Yang, Mingyu and Fu, Qiang and others},
  journal={arXiv preprint arXiv:2508.08601},
  year={2025}
}

@article{Motionstream,
  title={Motionstream: Real-time video generation with interactive motion controls},
  author={Shin, Joonghyuk and Li, Zhengqi and Zhang, Richard and Zhu, Jun-Yan and Park, Jaesik and Shechtman, Eli and Huang, Xun},
  journal={arXiv preprint arXiv:2511.01266},
  year={2025}
}

@inproceedings{Yume,
  title={Yume1. 5: A Text-Controlled Interactive World Generation Model},
  author={Mao, Xiaofeng and Li, Zhen and Li, Chuanhao and Xu, Xiaojie and Ying, Kaining and Zhang, Kaipeng},
  booktitle={Proceedings of the IEEE/CVF Conference on Computer Vision and Pattern Recognition},
  pages={7752--7761},
  year={2026}
}

@article{Magicworld,
  title={Magicworld: Interactive geometry-driven video world exploration},
  author={Li, Guangyuan and Zheng, Siming and Xu, Shuolin and Chen, Jinwei and Li, Bo and Hu, Xiaobin and Zhao, Lei and Jiang, Peng-Tao},
  journal={arXiv preprint arXiv:2511.18886},
  year={2025}
}

@article{InfiniteWorld,
  title={Infinite-World: Scaling Interactive World Models to 1000-Frame Horizons via Pose-Free Hierarchical Memory},
  author={Wu, Ruiqi and He, Xuanhua and Cheng, Meng and Yang, Tianyu and Zhang, Yong and Kang, Zhuoliang and Cai, Xunliang and Wei, Xiaoming and Guo, Chunle and Li, Chongyi and others},
  journal={arXiv preprint arXiv:2602.02393},
  year={2026}
}

@article{Worldplay,
  title={Worldplay: Towards long-term geometric consistency for real-time interactive world modeling},
  author={Sun, Wenqiang and Zhang, Haiyu and Wang, Haoyuan and Wu, Junta and Wang, Zehan and Wang, Zhenwei and Wang, Yunhong and Zhang, Jun and Wang, Tengfei and Guo, Chunchao},
  journal={arXiv preprint arXiv:2512.14614},
  year={2025}
}

@inproceedings{Gen3R,
  title={Gen3r: 3d scene generation meets feed-forward reconstruction},
  author={Huang, Jiaxin and Yang, Yuanbo and Yang, Bangbang and Ma, Lin and Ma, Yuewen and Liao, Yiyi},
  booktitle={Proceedings of the IEEE/CVF Conference on Computer Vision and Pattern Recognition},
  pages={25358--25369},
  year={2026}
}

@article{da3,
  title={Depth anything 3: Recovering the visual space from any views},
  author={Lin, Haotong and Chen, Sili and Liew, Junhao and Chen, Donny Y and Li, Zhenyu and Shi, Guang and Feng, Jiashi and Kang, Bingyi},
  journal={arXiv preprint arXiv:2511.10647},
  year={2025}
}

@article{Dens3r,
  title={Dens3r: A foundation model for 3d geometry prediction},
  author={Fang, Xianze and Gao, Jingnan and Wang, Zhe and Chen, Zhuo and Ren, Xingyu and Lyu, Jiangjing and Ren, Qiaomu and Yang, Zhonglei and Yang, Xiaokang and Yan, Yichao and others},
  journal={arXiv preprint arXiv:2507.16290},
  year={2025}
}

@article{Worldmirror,
  title={Worldmirror: Universal 3d world reconstruction with any-prior prompting},
  author={Liu, Yifan and Min, Zhiyuan and Wang, Zhenwei and Wu, Junta and Wang, Tengfei and Yuan, Yixuan and Luo, Yawei and Guo, Chunchao},
  journal={arXiv preprint arXiv:2510.10726},
  year={2025}
}

@article{meng20253d,
  title={3D indoor scene geometry estimation from a single omnidirectional image: A comprehensive survey},
  author={Meng, Ming and Zhu, Yonggui and Zhao, Yufei and Li, Zhaoxin and Zhu, Zhe},
  journal={Computational Visual Media},
  year={2025},
  publisher={TUP}
}

@article{wang2025diffusion,
  title={Diffusion models for 3D generation: A survey},
  author={Wang, Chen and Peng, Hao-Yang and Liu, Ying-Tian and Gu, Jiatao and Hu, Shi-Min},
  journal={Computational Visual Media},
  volume={11},
  number={1},
  pages={1--28},
  year={2025},
  publisher={TUP}
}

@article{wu2024recent,
  title={Recent advances in 3d gaussian splatting},
  author={Wu, Tong and Yuan, Yu-Jie and Zhang, Ling-Xiao and Yang, Jie and Cao, Yan-Pei and Yan, Ling-Qi and Gao, Lin},
  journal={Computational Visual Media},
  volume={10},
  number={4},
  pages={613--642},
  year={2024},
  publisher={TUP}
}

@article{peng2025gaussian,
  title={Gaussian-plus-SDF SLAM: High-fidelity 3D reconstruction at 150+ fps},
  author={Peng, Zhexi and Zhou, Kun and Shao, Tianjia},
  journal={Computational Visual Media},
  year={2025},
  publisher={TUP}
}

@article{zhou2025onevae,
  title={OneVAE: Joint Discrete and Continuous Optimization Helps Discrete Video VAE Train Better},
  author={Zhou, Yupeng and Li, Zhen and Ouyang, Ziheng and Chen, Yuming and Du, Ruoyi and Zhou, Daquan and Fu, Bin and Liu, Yihao and Gao, Peng and Cheng, Ming-Ming and others},
  journal={arXiv preprint arXiv:2508.09857},
  year={2025}
}
\end{document}